\documentclass[runningheads]{llncs}

\usepackage{eccv}   

\usepackage{eccvabbrv}

\usepackage{graphicx}
\usepackage{booktabs}

\usepackage{epsfig}
\usepackage{amsmath}
\usepackage{amssymb}
\usepackage{color}
\usepackage{makecell}
\usepackage{multirow}
\usepackage{bbding}

\usepackage{pifont}
\usepackage{xcolor}
\usepackage{comment}
\usepackage[table]{xcolor}

\usepackage{enumitem}
\usepackage{amssymb}

\usepackage{caption}

\usepackage[accsupp]{axessibility}  % Improves PDF readability for those with disabilities.

\definecolor{customlightblue}{RGB}{224, 235, 246}

\definecolor{blueLv5}{HTML}{90CAF9} % 最佳 (最深端已调浅)
\definecolor{blueLv4}{HTML}{BBDEFB}
\definecolor{blueLv3}{HTML}{E3F2FD}
\definecolor{blueLv2}{HTML}{F1F8FE}
\definecolor{blueLv1}{HTML}{F8FBFF} % 最浅 (接近白色)

\newcommand{\tablestyle}[2]{\setlength{\tabcolsep}{#1}\renewcommand{\arraystretch}{#2}\centering\small}
\newcommand{\xjqi}[1]{\textcolor[rgb]{1,0,0}{{[xjqi: #1]}}}

\makeatletter

\newcommand{\Rmnum}[1]{\expandafter\@slowromancap\romannumeral #1@}
\makeatother

\definecolor{myblue}{HTML}{3c59a4}
\usepackage[colorlinks,linktoc=all]{hyperref}
\usepackage[all]{hypcap}
\hypersetup{citecolor=myblue}
\hypersetup{linkcolor=red}

\usepackage{orcidlink}
\usepackage{marvosym}

\begin{document}

\title{OpenSpatial: A Principled Data Engine for Empowering Spatial Intelligence} 

\titlerunning{OpenSpatial}

\author{Jianhui Liu\inst{1,2}\thanks{Equal Contribution \quad \textsuperscript{\Letter} Corresponding Author \quad \textsuperscript{$\dagger$} Project Leader} \and
Haoze Sun\inst{1}\textsuperscript{*} \and
Wenbo Li\inst{1}\textsuperscript{\Letter}\textsuperscript{$\dagger$} \and
Yanbing Zhang\inst{1} \and
Rui Yang\inst{2} \and
Zhiliang Zhu\inst{1} \and
Yijun Yang\inst{1} \and
Shenghe Zheng\inst{1} \and
Nan Jiang\inst{1} \and
Jiaxiu Jiang\inst{1} \and
Haoyang Huang\inst{1} \and
Tien-Tsin Wong\inst{3} \and
Nan Duan\inst{1} \and
Xiaojuan Qi\inst{2}\textsuperscript{\Letter}}

\authorrunning{J. Liu et al.}

\institute{\textsuperscript{1}Joy Future Academy \quad 
% \textsuperscript{2}HKU \quad
% \textsuperscript{3}ECUST \quad
% \textsuperscript{4}HEU \quad
% \textsuperscript{5}HKUST(GZ) \quad
% \textsuperscript{6}HIT \quad
% \textsuperscript{7}PKU \quad
% \textsuperscript{8}Monash University \\
\textsuperscript{2}The University of Hong Kong \quad
\textsuperscript{3}Monash University \\
\email{jhliu0212@gmail.com}, \email{shz22@tsinghua.org.cn}, \email{fenglinglwb@gmail.com}\\
\url{https://github.com/VINHYU/OpenSpatial} }

\maketitle

% \begin{figure}[t]
% \begin{center}
%    \includegraphics[width=1.0\linewidth]{Fig/datapipline_cropped.pdf}
%    \caption{Illustration of the data engine. The left panel of the figure illustrates the data processing and annotation pipeline, while the right panel presents the detailed statistics of the dataset, comprising five key dimensions: Spatial Measurement (SM), Spatial Relationship (SR), Multi-view Consistency (MC), Camera Perception (CP), and Scene-Aware Reasoning (SAR).} 
%    \label{fig:datapipline}
% \end{center}
% \end{figure}

% \begin{figure}[t]
% \begin{center}
%    \includegraphics[width=1.0\linewidth]{Fig/data_case.png}
%    \caption{Overview of the OpenSpatial dataset. OpenSpatial-3M comprises 3M high-quality samples for spatial understanding, encompassing five primary categories: Spatial Measurement (SM), Spatial Relationship (SR), Multi-view Consistency (MC), Camera Perception (CP), and Scene-Aware Reasoning (SAR). Representative cases for each category are curated and illustrated above. For brevity, the displayed QA pairs have been condensed; please refer to the Appendix for comprehensive details.} 
%    \label{fig:LogiSpatial}
% \end{center}
% % \vspace{-0.3cm}
% \end{figure}

\begin{abstract}

Spatial understanding is a fundamental cornerstone of human-level intelligence. Nonetheless, current research predominantly focuses on domain-specific data production, leaving a critical void: the absence of a principled, open-source engine capable of fully unleashing the potential of high-quality spatial data. To bridge this gap, we elucidate the design principles of a robust data generation system and introduce OpenSpatial—an open-source data engine engineered for high quality, extensive scalability, broad task diversity, and optimized efficiency. OpenSpatial adopts 3D bounding boxes as the fundamental primitive to construct a comprehensive data hierarchy across five foundational tasks: Spatial Measurement (SM), Spatial Relationship (SR), Camera Perception (CP), Multi-view Consistency (MC), and Scene-Aware Reasoning (SAR). Leveraging this scalable infrastructure, we curate OpenSpatial-3M, a large-scale dataset comprising 3 million high-fidelity samples. Extensive evaluations demonstrate that versatile models trained on our dataset achieve state-of-the-art performance across a wide spectrum of spatial reasoning benchmarks. Notably, the best-performing model exhibits a substantial average improvement of 19\%, relatively. Furthermore, we provide a systematic analysis of how data attributes influence spatial perception. By open-sourcing both the engine and the 3M-scale dataset, we provide a robust foundation to accelerate future research in spatial intelligence.
\end{abstract}

\begin{figure}[t]
\begin{center}
   \includegraphics[width=1.0\linewidth]{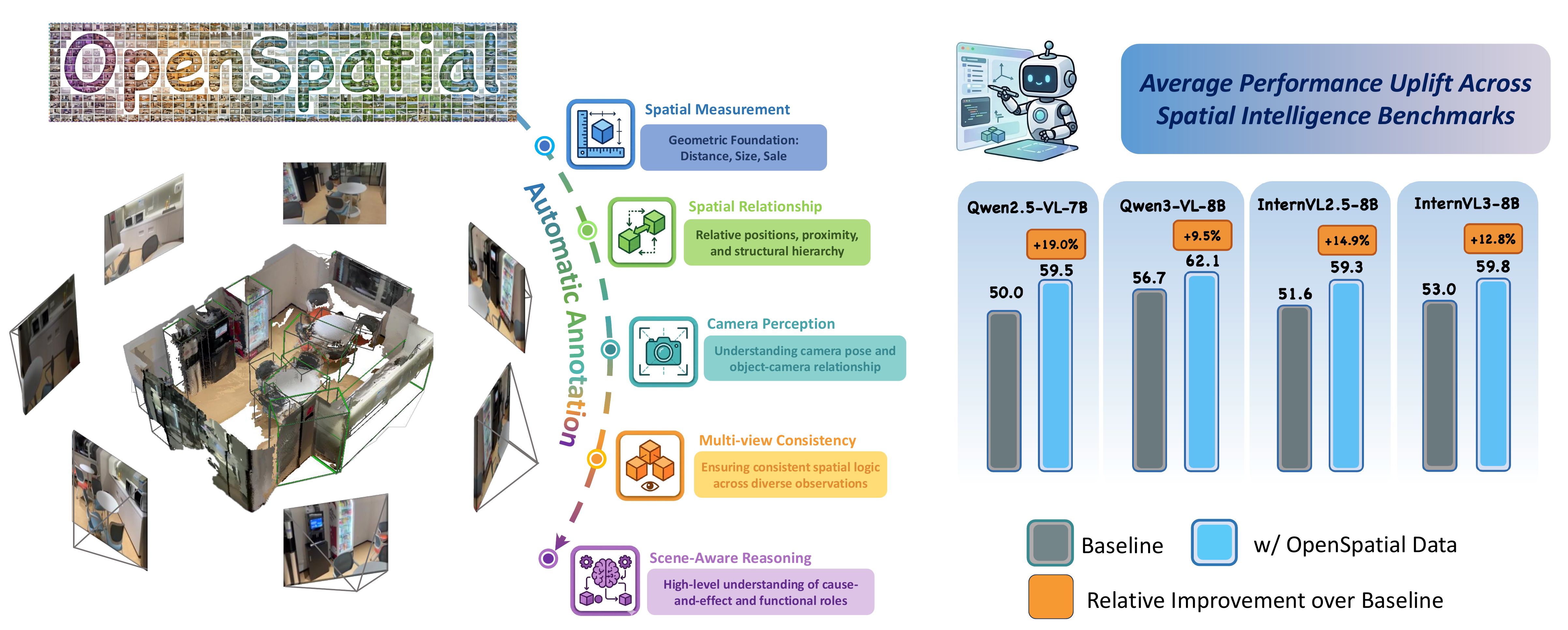}
   \caption{Overview of OpenSpatial. The left panel provides a high-level schematic of the OpenSpatial pipeline. The right panel demonstrates that models trained with OpenSpatial-generated data exhibit significant improvements in spatial intelligence. Notably, the evaluation benchmarks are consistent with those reported in Tab.~\ref{tab:main-results}.} 
   \label{fig:teaser}
\end{center}
\vspace{-1.0cm}
\end{figure}

\section{Introduction}
\label{sec:intro}

Multi-modal large language models (MLLMs)~\cite{seed1.5, seed1.8, qwen3-t, gemini2.5, grok15v2024, gpt4, kimi, team2025kimi-vl} have progressed from image-text alignment to instruction-following systems capable of visual intelligence. Yet their spatial competence still trails their semantic expressiveness: models can produce convincing descriptions but often fail to perceive accurate distance, maintain multi-view consistency, or construct spatial cognitive maps -- capabilities central to embodied decision-making~\cite{dynscene, bip3d} and robotics~\cite{rt, goyal2023rvt, shridhar2023perceiver}. This gap has motivated ``spatialized'' VLMs~\cite{3Dthink, cambrian-s, VST, vlm-3r, spacer} and dedicated benchmarks~\cite{blink, allangles, vsi, mmsi}, yielding measurable improvements in spatial understanding. However, these gains remain uneven across tasks and scenes, suggesting that the bottleneck is not architecture alone but the foundations of spatial generalization.

A major foundation is \textbf{data}: which spatial signals are present, how they are synthesized, and what distributions they cover. Current data-centric efforts, however, face two systemic obstacles. First, the limited diversity of current spatial data constrains the robustness of state-of-the-art models. This data scarcity leads to ``spatial myopia'', where models exhibit high benchmark scores but lack the versatility required for real-world environments. Second, and more critically, much spatial data is generated by closed, under-specified pipelines. Existing works~\cite{VST, cambrian-s, sensenova-si} release only fixed, preprocessed datasets (sometimes in limited subsets) while keeping their generative engines proprietary, making it difficult to run controlled ablations, scale data in a consistent way, or study which design choices actually drive spatial capability. This black-box data ecosystem fragments progress into disconnected silos and raises the barrier to reproducible, systematic advancement.

%Recognizing these limitations, recent research has increasingly turned toward data-centric strategies to augment the spatial reasoning of MLLMs. However, these efforts are fundamentally constrained by two systemic bottlenecks. First, the diversity of spatial annotation remains severely impoverished. While pioneering models such as SenseNova-SI~\cite{sensenova-si} or Cambrian-S~\cite{cambrian-s} have achieved competitive results on specific benchmarks, their reliance on narrow data distributions often leads to "spatial myopia"—a failure to generalize beyond limited task types. This lack of diverse, high-granularity spatial cues prevents models from internalizing a holistic understanding of 3D environments. Second, and more critically, the community suffers from a pervasive "black-box" approach to data generation. Researchers frequently release only static, pre-processed datasets or their minimal subsets, while keeping the underlying generative engines proprietary. This practice traps the field in a cycle of fragmented data silos, where the lack of scalable, consistent, and open-sourced infrastructure stifles systematic exploration into the scaling laws of spatial perception. Consequently, this imposes a formidable barrier to entry, fragmenting research efforts and significantly curtailing the collective evolution of embodied spatial intelligence. 
We argue that advancing spatial intelligence requires moving beyond static dataset releases toward open, reusable data infrastructure. We therefore introduce \textbf{OpenSpatial}, as shown in Fig.~\ref{fig:teaser}, an open-source data engine that synthesizes high-quality, scalable, and task-diverse supervision for spatial understanding. OpenSpatial is built on three key designs. \textbf{(1) 3D box–centric grounding for quality}: by anchoring supervision in object-aligned 3D boxes rather than 2D projections~\cite{hartley2003multiple}, it yields high-fidelity, viewpoint-consistent labels that capture true 3D structure and support metric reasoning. \textbf{(2) 3D lifting for scalability}: it automatically elevates sparse cues into high-quality 3D box priors, enabling data generation to extend beyond curated sets to unconstrained, in-the-wild sources. \textbf{(3) Scene-graph-driven synthesis for diversity}: it programmatically enumerates objects, attributes, and relations to generate balanced QA across measurement, relations, camera changes, multi-view consistency, and scene-level reasoning, mitigating ``spatial myopia''. OpenSpatial is further engineered for throughput via parallel execution and feature reuse, enabling rapid large-scale data annotations. Together, these choices make spatial supervisions transparent and controllable, supporting principled ablations, reliable scaling, and improved generalization. 

Built on this infrastructure, we curate OpenSpatial-3M, a 3-million-sample training suite spanning five core capabilities-- Spatial Measurement, Spatial Relationship, Camera Perception, Multi-view Consistency, and Scene-Aware Reasoning, organized as a progressive curriculum that bridges egocentric observations with stable world-coordinate understanding. We show that models fine-tuned on the data achieve state-of-the-art performance on challenging spatial benchmarks (\textit{e.g.}, BLINK, AllAngles, MMSI), consistently surpassing strong open-source baselines with large average gains. As shown in Fig.~\ref{fig:teaser}, the OpenSpatial-3M dataset consistently enhances performance across various architectures, yielding a \textbf{14.1\%} average improvement and a maximum gain of \textbf{19\%} over the baseline.. Beyond performance improvements, OpenSpatial’s modular pipeline enables controlled analyses of which design choices drive improvements (\textit{e.g.}, box-centric grounding and data filtering), supporting reproducible, data-driven scaling of spatial perception across architectures.

In summary, this work advances the frontier of spatial intelligence through three key contributions.
\begin{itemize}
\item We introduce \textbf{OpenSpatial}, an open-source, controllable data engine for synthesizing high-quality, scalable, and task-diverse spatial supervision.
\item We release \textbf{OpenSpatial-3M}, a large-scale curriculum-style training suite covering five foundational spatial capabilities.
\item We demonstrate strong empirical gains and provide diagnostic analyses enabled by the engine’s modularity, clarifying how specific data designs improve spatial generalization and offering a reproducible foundation for future work.
  %\item We present OpenSpatial, a principled, open-source data engine that establishes a high-qulity infrastructure for spatial reasoning data synthesis.
  %\item We introduce OpenSpatial-3M, a massive dataset with 3 million diverse spatial understanding samples structured across a hierarchical task curriculum.
  %\item Extensive experiments across multiple MLLM architectures, we demonstrate that OpenSpatial-3M consistently achieves state-of-the-art performance on major benchmarks, significantly outperforming existing open-source models.
\end{itemize}

\section{Related Work}

\vspace{0.05in} \noindent\textbf{Large Vision-Language Models.}
The rapid evolution of Large Vision-Language Models (LVLMs) has revolutionized multimodal intelligence by bridging high-dimensional visual perception with complex linguistic reasoning, a progress fundamentally anchored in the advancement of Visual Instruction Tuning. This paradigm, pioneered by LLaVA~\cite{llava}, demonstrated that projecting deep visual features into the LLM’s embedding space via a lightweight interface enables the model to follow complex multimodal prompts with sophisticated cross-modal logic. Building upon this foundation, the Qwen series~\cite{qwen, qwen2, qwen3, qwen3-t} introduced critical architectural refinements—specifically the Naive Dynamic Resolution mechanism for processing visual inputs of arbitrary aspect ratios and the transition toward a unified multimodal backbone—significantly bolstering fine-grained comprehension and complex scene analysis. Modern LVLMs typically employ a hierarchical training pipeline, progressing from an initial feature alignment phase that synchronizes visual tokens with linguistic semantics to extensive Supervised Fine-Tuning (SFT) on diverse, instruction-rich datasets. As exemplified by the scaling strategies of InternVL~\cite{internvl, internvl3, internvl3.5} and the query-based alignment in InstructBLIP~\cite{instructblip}, this trajectory has shifted the field from rudimentary image-text matching toward robust visual intelligence. 
% Furthermore, the incorporation of advanced Reinforcement Learning (RL) frameworks and Chain-of-Thought (CoT)~\cite{cot} reasoning has further optimized the models' alignment with complex human instructions, fostering superior decision-making and logical consistency in increasingly intricate and long-context real-world environments.

\vspace{0.05in} \noindent\textbf{Large Vision-Language Models for Spatial Reasoning} Despite the significant strides in general image and video understanding, existing LVLMs still struggle with sophisticated spatial reasoning tasks that necessitate the interpretation of intricate geometric transformations and spatial configurations. To enhance the spatial intelligence of Large Vision-Language Models (LVLMs), existing research has diverged into architectural augmentation, large-scale dataset curation~\cite{mmspatial, multi-spatialmllm, osworld}, and advanced training paradigms~\cite{long2026spatialreward}. Architecturally, models such as Spatial-MLLM~\cite{spatialmllm}, VLM-3R~\cite{vlm-3r}, and 3DThinker~\cite{3Dthink} incorporate geometric priors via external 3D encoders, while SpatialBot~\cite{spatialbot} and VILASR~\cite{wu2025reinforcing} utilize external tools for  depth estimation and ground perception. Simultaneously, the field has transitioned toward data-driven scaling; SpatialVLM~\cite{spatial-vlm} and SpatialRGPT~\cite{spatialrgpt} pioneered the synthesis of massive spatial VQA datasets, a trajectory further extended by VST~\cite{VST} and SenseNova-SI~\cite{sensenova-si}, and they bolster the model's comprehension capabilities by scaling up both the volume and the diversity of the training data. To refine reasoning capabilities, multi-stage frameworks like SpatialLadder~\cite{spatialladder} and Cambrian-S~\cite{cambrian-s} employ progressive SFT, whereas SpaceR~\cite{spacer} and MindCube~\cite{mindcube} integrate cognitive maps with reinforcement learning to optimize reasoning traces.  While these methodologies have introduced various specialized datasets, they often lack a granular analysis from a data-centric perspective on how specific construction strategies fundamentally enhance a model's spatial intelligence. To bridge this gap, we systematically investigate the principles of spatial data synthesis and introduce a more curated, large-scale dataset that specifically addresses previously overlooked perspective-taking tasks, thereby fostering a more holistic spatial reasoning framework.

\begin{figure}[t]
\begin{center}
   \includegraphics[width=1.0\linewidth]{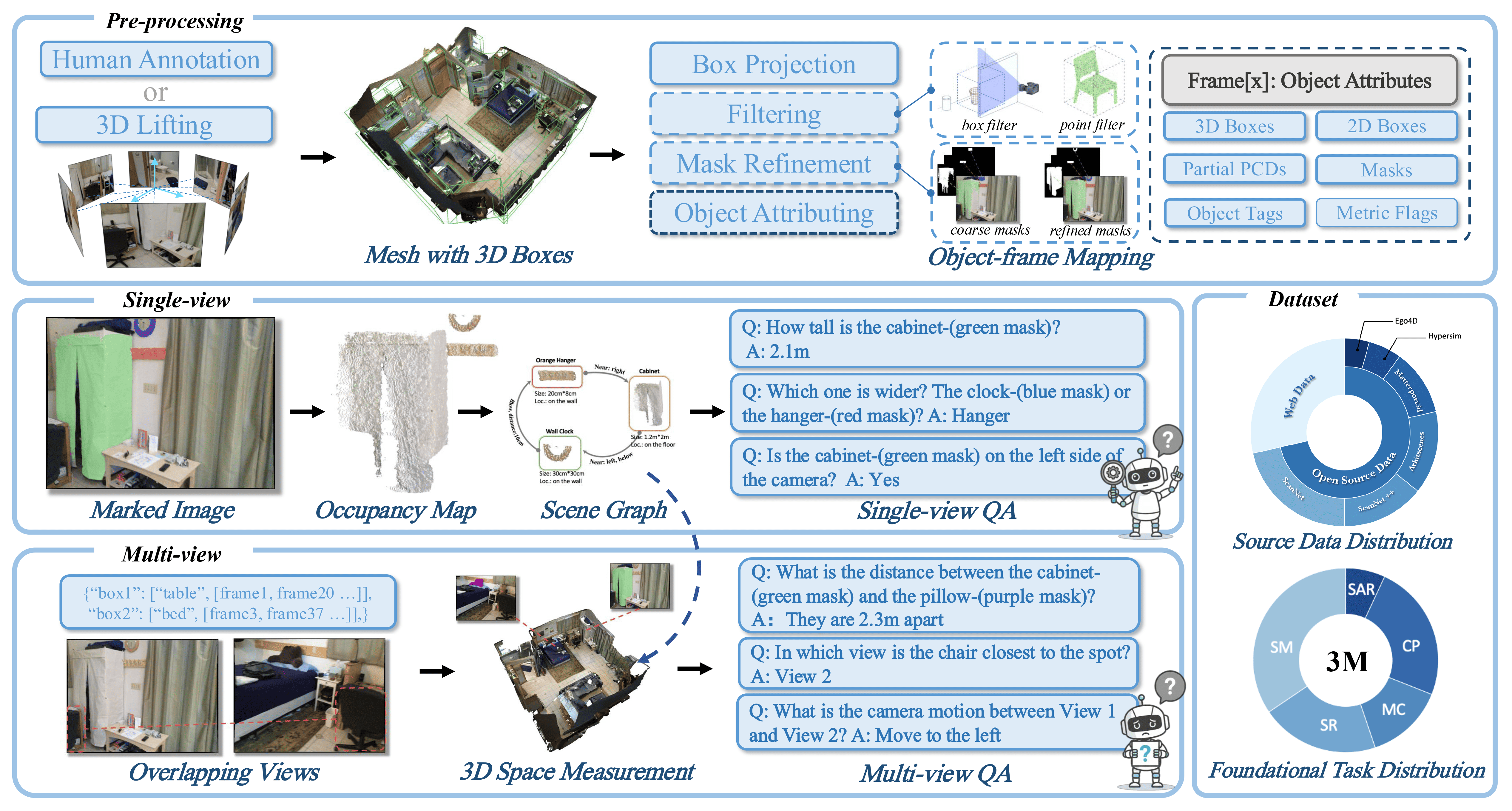}
   \caption{Illustration of the data engine. The left panel of the figure illustrates the data processing and annotation pipeline, while the right panel presents the detailed statistics of the dataset, including source data distribution and task distribution.} 
   \label{fig:datapipline}
\end{center}
\vspace{-1.0cm}
\end{figure}

% \section{OpenSpatial: An Open-source Data Engine for Spatial Intelligence} 
\section{Implementation Principles of OpenSpatial} 

\begin{figure}[t!]
\begin{center}
   \includegraphics[width=1.0\linewidth]{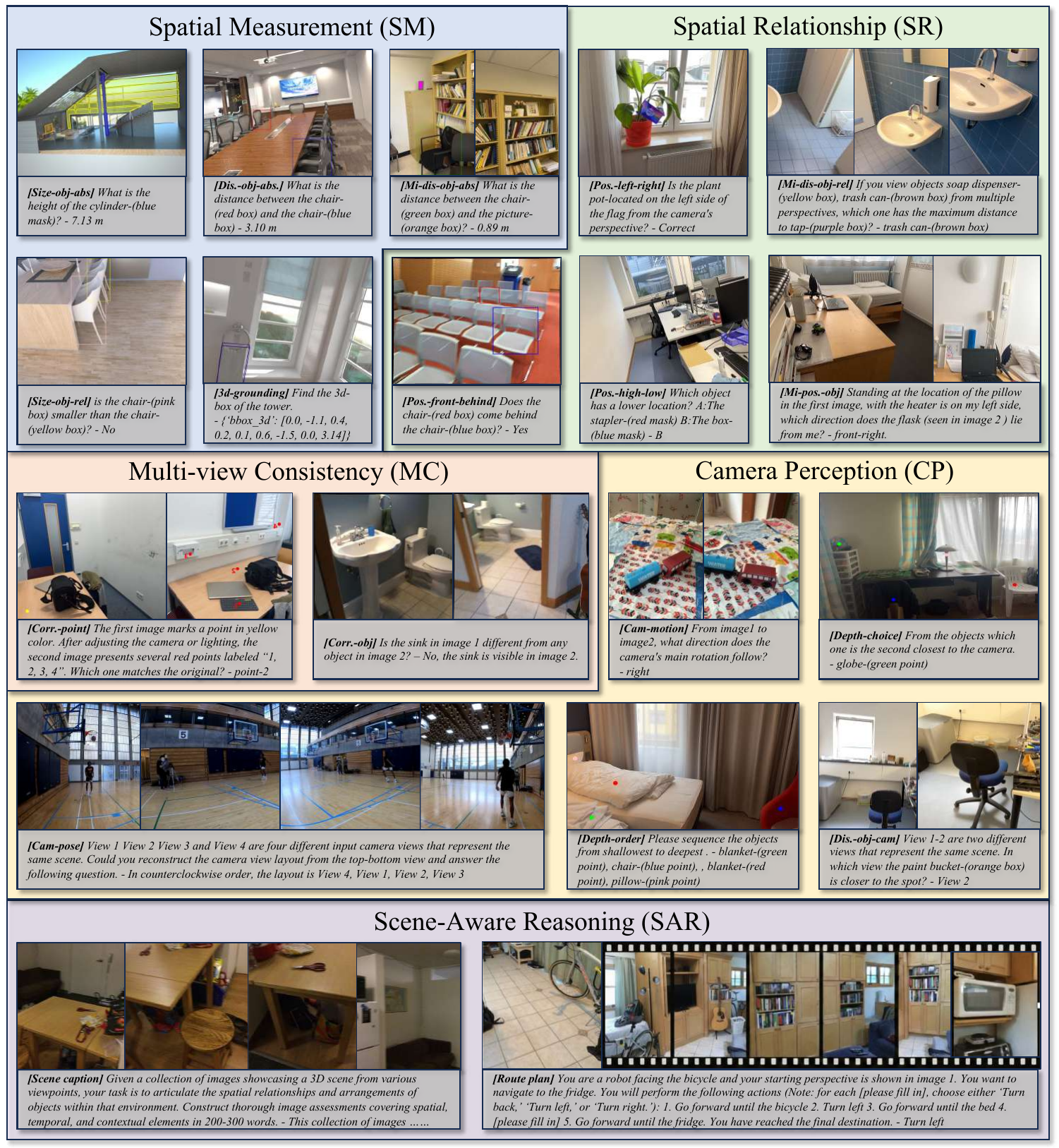}
   \caption{Overview of the OpenSpatial dataset. OpenSpatial-3M comprises 3M high-quality samples for spatial understanding, encompassing five primary categories: Spatial Measurement (SM), Spatial Relationship (SR), Camera Perception (CP), Multi-view Consistency (MC),  and Scene-Aware Reasoning (SAR). Representative cases for each category are curated and illustrated above. For brevity, the displayed QA pairs have been condensed; please refer to the Appendix for comprehensive details.} 
   \label{fig:LogiSpatial}
\end{center}
\vspace{-0.7cm}
\end{figure}

To enable reproducible progress in spatial intelligence, we introduce \textbf{OpenSpatial}, an open-source data engine that generates spatial supervision from a unified \emph{3D box–centric} representation (Fig. \ref{fig:datapipline}). Unlike static dataset releases, OpenSpatial exposes the full data-production pipeline and supports two complementary annotation modes: (i) \emph{human annotation} for maximal accuracy, and (ii) \emph{automated 3D lifting} for scalable expansion to in-the-wild web data and open-source assets. Both modes output the same canonical format-- a scene mesh with object-aligned 3D boxes-- which then drives consistent attribute extraction and QA synthesis. Built on this engine, we curate \textbf{OpenSpatial-3M}, a 3M-entry training suite covering five foundational capabilities (SM/SR/CP/MC/SAR in Fig. \ref{fig:LogiSpatial}), further divided into 19 sub-tasks, forming a scalable and extensible foundation for general-purpose spatial understanding.

%To foster community growth and provide a robust foundation for spatial intelligence, OpenSpatial is presented as an open-source, comprehensive data engine centered around 3D bounding boxes as its core representation. OpenSpatial breaks through the bottlenecks of traditional data production by offering versatile annotation modes: it supports fully automated annotation pipelines while also providing the flexibility for selective human intervention. This dual-capability ensures the rapid scaling up of data production while maintaining exceptional diversity, high precision, and efficiency. Leveraging this robust engine, we have curated the OpenSpatial-3M dataset, comprising 3 million high-quality spatial understanding entries. This dataset encompasses five foundational spatial tasks, which are further categorized into 19 specialized sub-tasks, establishing a solid data framework for advancing general-purpose spatial perception.

\subsection{Data Pipeline}
OpenSpatial turns raw multi-view images (or video keyframes) into spatial question-answer pair through a staged pipeline (Fig. \ref{fig:datapipline}). It starts by producing scene-level 3D oriented bounding boxes (OBBs) for objects, either via manual labeling or via an automated lifting procedure. These scene-level boxes are then converted into frame-level object attributes through projection, visibility filtering, and mask refinement, resulting in a consistent object–frame index (3D/2D boxes, masks, partial point clouds, tags, and metric flags). This shared representation supports two downstream annotation branches: single-view QA, generated from per-frame scene graphs with explicit visual anchors, and multi-view QA, generated by sampling overlapping views and using the viewpoint-invariant 3D boxes to enforce cross-view correspondence and consistency.

\vspace{0.05in}\noindent\textbf{3D Box-Centric Design:} 
Spatial understanding requires a stable 3D scene state: object position, size, orientation, and relations under viewpoint changes and occlusion. OpenSpatial uses Oriented Bounding Boxes (OBBs) as the core representation because they offer a compact, scalable middle ground between weak 2D labels and expensive dense 3D reconstructions. OBBs are world-coordinate and viewpoint-invariant, giving each object a single consistent 3D anchor across frames, which makes cross-view association and supervision straightforward. They also encode the minimum 3D structure needed for spatial reasoning (depth, extent, orientation), supporting metric, topological, and directional relations. Finally, OBBs act as a canonical anchor for downstream processing, enabling consistent projection, visibility/occlusion filtering, and box-conditioned mask refinement to synchronize 3D geometry with precise 2D visual grounding.  Concretely, each object is parameterized as an OBB $(x, y, z, x_l, y_l, z_l, r, p, y)$: $(x, y, z)$ is the box center in world coordinates, $(x_l, y_l, z_l)$ are the side length along the X/Y/Z axes, and $(r, p, y)$ are Roll/Pitch/Yaw. All boxes are defined in a global world coordinate system with a Z-up convention, providing a consistent geometric anchor shared across frames and camera trajectories.   

%Defined as the ability to perceive and reason about object locations, orientations, and spatial interrelationships, spatial understanding necessitates high-fidelity object modeling. In our data engine, we adopt Oriented Bounding Boxes (OBBs) to serve as a robust 3D representation for each object. The inherent advantage of OBBs lies in their ability to encode intrinsic 3D properties while remaining invariant to changes in camera viewpoints. Specifically, the 3D bounding box is defined by $(x, y, z, x_l, y_l, z_l, r, p, y)$, where $(x, y, z)$ are the center of the box, $(x_l, y_l, z_l)$ are the length along the X, Y, and Z axes, and (r, p, y) are the rotation angles of Roll, Pitch, Yaw. For each annotated scene, the OBBs are defined within the world coordinate system, adhering to a Z-axis-up convention. 

\vspace{0.05in}\noindent\textbf{Scene-level Bounding Box Annotation:}  Given a posed video sequence, our first step is to obtain an oriented 3D bounding box for each object. OpenSpatial supports two complementary annotation modes. \emph{Manual annotation}, following the EmbodiedScan protocol~\cite{embodiedscan}, leverages human effort to label objects in 3D and yields high-precision boxes, but is time-consuming and difficult to scale. To extend beyond curated datasets to web data and open-source assets without fine-grained labels, we additionally provide an \emph{automated 3D lifting} pipeline. Starting from video keyframes or multi-view images, we perform per-view object recognition with Gemini~\cite{gemini2.5} and instance mask extraction with SAM~\cite{SAM}, then associate and merge instances in 3D space and fit a convex hull to produce the final oriented boxes. Qualitative visualizations are reported in the experiments (Fig.~\ref{fig:3d-lifting}). After this step, each scene is represented as a reconstructed 3D mesh together with a set of object-aligned oriented 3D bounding boxes (Fig.~\ref{fig:datapipline}). 

%Our data engine supports two distinct modes for scene-level bounding box annotation: first, it supports high-quality manual annotation (similar to the meticulous labeling found in EmbodiedScan)~\cite{embodiedscan}, which provides superior precision but is relatively time-consuming; second, to effectively scale up to web data or open-source datasets lacking fine-grained labels, we employ an automated 3D lifting strategy. This automated pipeline reconstructs 3D instances from video keyframes or multi-view images by leveraging Gemini~\cite{gemini2.5} for per-view object recognition and SAM~\cite{SAM} for mask generation, followed by a semantic merge in 3D space and convex hull calculation to derive the final 3D oriented bounding boxes. Visualizations of these results are provided in the experimental section.

\vspace{0.05in}\noindent\textbf{Attribute-Centric Object–Frame Mapping.} 
After obtaining a scene mesh with object-aligned 3D boxes, the next step is to convert these \emph{scene-level} annotations into \emph{frame-level} attributes that can reliably support downstream scene-graph construction and QA synthesis across diverse task types.  To this end, our engine extracts a comprehensive set of object attributes from the initial 3D bounding boxes (top-right part of Fig.~\ref{fig:datapipline}). We first project each 3D box onto individual frames, then apply two filters to ensure data integrity: \textbf{(1)} boxes outside the current camera frustum are discarded; \textbf{(2)} to handle occlusions where an object projects into the frame but is invisible or heavily truncated, we perform depth-based validation. Specifically, pixels within the projected 2D box are back-projected into world coordinates using the depth map and camera intrinsics/extrinsics to form a local point cloud, and we compute the volumetric occupancy of these points inside the 3D box. Boxes with occupancy below a threshold are removed as occluded.

For boxes that pass filtering, the validated point-cloud pixels define coarse masks, which are further refined by SAM~\cite{SAM} to obtain fine-grained 2D instance masks that tightly align 3D geometry with visual appearance. Because automated labeling may contain duplicate objects with similar semantics, these masks serve as robust instance indicators and can be further converted into bounding boxes or keypoints as precise spatial prompts. Finally, we consolidate all extracted attributes-- including masks, 2D/3D boxes, partial point clouds, and object tags-- into a structured indexing system across frames. Each object is also assigned a \emph{metric flag} indicating whether its box reflects real-world scale; if False, measurement-related QA generation is skipped to avoid noisy supervision. This object--frame indexing forms a unified and reliable foundation for all subsequent single-view and multi-view spatial understanding tasks.

\vspace{0.05in}\noindent\textbf{Scene-Graph-Driven QA Synthesis for Diverse Spatial Supervision}  Given the object-frame indexing produced in the previous step, OpenSpatial synthesizes spatial QA in two complementary settings: single-view and multi-view-- with an explicit emphasis on task diversity. The engine programmatically enumerates objects, their attributes, and inter-object relations (via scene graphs) to generate a balanced collection of questions spanning measurement, spatial relations, camera/viewpoint changes, multi-view consistency, and scene-level reasoning, thereby mitigating the narrow coverage that often leads to ``spatial myopia''. Concretely, we generate QA in the following two types: 

\begin{itemize}

\item \textit{Single-view annotation.} For each frame, we construct a structured scene graph from the indexed objects and attributes (e.g., 2D/3D boxes, masks, tags). To prevent referential ambiguity when multiple instances share similar semantics, we render a marked image that highlights the queried object(s) as explicit anchors. Conditioned on the marked image and the scene graph, we generate diverse single-view QA that probes object–object and object–environment relationships, including relational queries (\textit{e.g.}, left, right, front, behind), attribute comparisons (\textit{e.g.}, size, relative depth), and context-dependent reasoning grounded in the current view.

\item \textit{Multi-view annotation}. Multi-view QA targets cross-view spatial reasoning, requiring consistent correspondence and geometry under viewpoint changes. The key challenge is selecting view pairs with sufficient overlap to enable inference while still introducing meaningful viewpoint variation. Our 3D box–centric representation provides a principled solution: because 3D boxes are anchored in a global world coordinate system, they serve as viewpoint-invariant references for linking instances across frames. We therefore sample view pairs that share a subset of 3D boxes, ensuring both contextual overlap and diversity. For each pair, we build a unified multi-view scene graph that merges object instances across views and generate cross-view QA-- such as re-identification under perspective shifts, reasoning about camera changes, and consistency or measurement checks when permitted-- encouraging the model to form a coherent 3D representation that generalizes across viewpoints.

\end{itemize}

\subsection{Description of OpenSpatial-3M Dataset}

\vspace{0.05in}\noindent\textbf{Data Source}: Following VST~\cite{VST}, we leverage the meticulously annotated 3D bounding boxes from EmbodiedScan~\cite{embodiedscan} as the foundational data for our pipeline, which aggregates scenes from ScanNet~\cite{scannet}, Matterport3D~\cite{matterport3d}, ARKitScenes~\cite{arkitscenes}, and SUN-RGBD~\cite{sunrgbd}. Notably, we excluded SUN-RGBD due to its relatively lower annotation fidelity. To further enhance environmental diversity, we incorporated pre-processed data from ScanNet++~\cite{scannet++} and Hypersim~\cite{hypersim} as supplementary sources. Additionally, we collected a set of web data to further broaden the diversity of our data sources. By including these real-world images/videos, we improve the dataset's coverage of various scenarios, ensuring better generalization across different environments.

\vspace{0.05in}\noindent\textbf{Task Taxonomy}: We decompose spatial understanding into five core capabilities, as shown in Fig.~\ref{fig:LogiSpatial}, each further categorized into a diverse set of sub-tasks. A detailed characterization of these dimensions is provided in the following:
\begin{itemize}
\item {Spatial Measurement (SM):} Spatial measurement is the process of quantifying the geometric metrics of objects and their configurations within a 3D coordinate system. It involves estimating absolute physical scales, such as length, width, height, and distance, serving as the foundational building blocks of spatial understanding.
\item{Spatial Relationship (SR):} Spatial Relationship characterizes the 3D spatial arrangement between entities, focusing on relative localization and inter-object dependencies. It provides the qualitative framework necessary to describe the layout of a scene beyond individual object coordinates.
\item{Camera Perception (CP):} Camera Perception helps model estimate camera poses and relative object-camera relationship. This sensor-aware intelligence serves as a foundational prior for implicit 3D reconstruction, enabling the translation of 2D observations into structured 3D coordinate systems.
\item{Multi-view Consistency (MC):} Multi-view Consistency serves as the cornerstone for scene-level understanding. It aims to establish robust spatial correspondences across diverse viewpoints by identifying shared objects and environmental contexts. By correlating the same physical entities from different perspectives, this capability requires the model to maintain a persistent 3D representation, ensuring that spatial reasoning remains coherent despite changes in camera pose.
\item{Scene-Aware Reasoning (SAR):} Scene-Aware Reasoning focuses on holistic scene-level understanding and long-range spatial logic. It empowers the model with the ability to perceive spatial layouts and perform  planning or navigation. By synthesizing the spatial configuration of obstacles and open spaces, the model develops the high-level reasoning required to determine traversability and optimal movement within a complex 3D environment.
\end{itemize}

\section{Experiments}

\subsection{Implementation Details}
\subsubsection{Training Setting:} We perform supervised fine-tuning (SFT) on several representative open-source VLMs to evaluate the effectiveness of our data engine. Adhering to the training protocol established in VST~\cite{VST}, each model is trained for a single epoch using 32 NVIDIA GPUs with a global batch size of 128, unless specified otherwise. We employ the AdamW optimizer for parameter updates, setting the base learning rate to $5 \times 10^{-5}$. We apply a decoupled learning rate strategy, setting the vision encoder's learning rate to a smaller $5 \times 10^{-6}$. 

\subsubsection{Training Data:} In our experiments, we primarily utilize the OpenSpatial-3M dataset. To further extend our coverage and incorporate diverse spatial scenarios, we integrate the high-quality open-source dataset SenseNova-800K into our training mixture, supplementing spatial reasoning dimensions not fully addressed by our primary corpus.  Furthermore, to maintain the models' general multimodal capabilities while enhancing their spatial intelligence, we employ a 1:1 ratio of general multi-modal data from LLaVA-OneVision~\cite{llava-ov} and spatial reasoning data from OpenSpatial. 

\subsubsection{Evaluation:} We evaluate the spatial reasoning capability of the VLMs across several representative benchmarks: BLINK~\cite{blink}, AllAngles~\cite{allangles}, ERQA~\cite{erqa}, VSI~\cite{vsi}, 3D-SR~\cite{3dsrbench}, MMSI~\cite{mmsi}, CVBench-3D~\cite{cv-bench}, and RealWorldQA~\cite{grok15v2024}. For general multimodal capabilities, we extend our evaluation to the MMStar~\cite{mmstar}, MMBench~\cite{mmbench}, and MMMU~\cite{mmmu} benchmarks. All models are assessed under identical settings using their respective native system prompts to ensure a fair comparison and minimize the impact of prompt variability.

\begin{table*}[t]
\centering
% \scalebox{0.9}{ %此处放置命令
\caption{Performance comparison of representative VLMs on spatial reasoning and general multimodal benchmarks. \colorbox{blueLv4}{Dark blue} and \colorbox{blueLv3}{light blue} backgrounds denote the best and second-best results, respectively. All models are evaluated under the same codebase to ensure a fair comparison. \textbf{Note}: 1)VSI is tested with 16 frames. 2)`$\pm$' denotes the performance gain or loss relative to the baseline model.}\label{tab:main-results}
\vspace{-0.3cm}

\resizebox{\linewidth}{!}{
\tablestyle{0.8pt}{1.0}
\begin{tabular}{l | c | c c c c c c c c | c c c c}
% \Xhline{0.5 pt}
\toprule
\textbf{Models} & \textbf{3D-Avg} & \textbf{BLINK} & \textbf{AllAngles} & \textbf{ERQA} & \textbf{VSI} & \textbf{3DSR} & \textbf{MMSI} & \textbf{CV-3D} & \textbf{RealWorldQA} & \textbf{MMStar} & \textbf{MMB} & \textbf{MMU} \\
% \Xhline{0.3 pt}
\midrule

\textbf{\textit{Proprietary Models}} &  &  &  &  &  &  &  &  & & & & \\
Gemini-2.5-Pro~\cite{gemini2.5} & 62.4 & 70.6 & 61.3 & 55.8 & 48.4 & 57.6 & 36.9 & 91.3 & 77.3 & 77.5 & 90.1 & 81.7 \\

\midrule

\textbf{\textit{Open-source General Models}} &  &  &  &  &  &  &  &  & & & & \\
% deepseek-VL-7B &  & 41.1 & 43.6 & 29.8 & 0.0 & 45.0 & 27.7 & 63.5 & 53.6 & 41.2 &  & 36.8 \\
InternVL2.5-4B~\cite{internvl3} & 47.3 & 50.8 & 45.1 & 41.0 & 28.3 & 44.0 & 28.5  & 76.4 & 64.2 & 58.5 & 78.5 & 50.0\\
InternVL2.5-8B~\cite{internvl3} & 51.6  & 54.9 & 48.9 & 40.8 & 39.3 & 51.0 & 28.6 & 79.9 & 69.4 & 62.6 & 82.3 & 53.3\\

InternVL3-2B~\cite{internvl3} & 47.9 & 52.8 & 48.6 & 36.3 & 30.3 & 46.4 & 25.9 & 77.3 & 65.5 & 61.5 & 77.7 & 45.9 \\
InternVL3-8B~\cite{internvl3} & 53.2 & 55.7 & 50.5 & 40.5 & 38.7 & 52.7 & 30.9 & 86.0 & 70.6 & 68.5 & 82.0 & 57.7 \\
Qwen2.5-VL-3B-Instruct~\cite{qwen2} & 45.6 & 49.0 & 42.8 & 40.8 & 32.0 & 45.2 &  25.0  & 64.8 & 65.2 & 56.6 & 77.2 & 48.4 \\
Qwen2.5-VL-7B-Instruct~\cite{qwen2} & 50.0 & 55.3 & 50.1 & 41.0 & 36.0 & 49.0 & 26.5 & 73.8 & 68.1 & 65.3 & 82.3 & 55.2 \\
Qwen3-VL-4B-Instruct~\cite{qwen3} & 56.2 & 62.6 & 49.1 & 40.2 & 53.6 & 52.5 & 28.0 & 92.3 & 71.4 & 67.5 & 82.8 & 57.7 \\
Qwen3-VL-8B-Instruct~\cite{qwen3} & 56.7 & 66.1 & 49.5 & 40.1 & 55.6 & 52.8 & 28.1 & 90.8 & 70.7 & 70.1 & 83.8 & 60.2 \\
Deepseek-VL2-27B-A4.5~\cite{deepseek-vl2} & - & 54.3 & 46.2 & 40.8 & - & 50.1 & 29.0 & 79.1 & 70.2 & 62.3 & 81.3 & 51.3 \\
MIMO-VL-7B-SFT~\cite{mimo-vl} & 54.6 & 59.7 & 52.9 & 41.0 & 37.5 & 56.1 & 29.3 & 86.9 & 73.5 & 71.1 & 80.9 & 65.9 \\
% \Xhline{0.3 pt}
\midrule
\textbf{\textit{Open-source SI Models}} &   &  &  &  &   &  &  &  & & & & \\
% LLava-OneVision-7B~\cite{llava-ov} & -  &  48.2 & - & - & 32.4  & - & 26.6 & - & 66.3 & 61.7 & 80.8 & 48.8 \\
SpaceR-7B~\cite{spacer} &  50.8 & 54.3 & 49.8 & 40.5 & 44.4  & 47.5 & 29.4 & 76.3 & 64.2 & 63.9 & 81.2 & 54.3 \\
SenseNova-SI-1.1-Qwen2.5-VL-7B~\cite{sensenova-si} & 51.8 & 55.0 & 47.7 & 39.0 & 55.2 & 46.7 & 32.5 & 77.4 & 60.5 & 58.9 & 80.4 & 48.0 \\
SenseNova-SI-1.1-Qwen3-VL-8B~\cite{sensenova-si} & 55.5  & 56.4 & 47.3 & 41.8 & 58.8  & 51.8 & 34.6 & 89.2 & 63.8 & 65.0 & 80.6  & 59.8 \\

VST-7B-SFT~\cite{VST} & 57.9 &  62.1 & 49.5 & 43.8 & 55.3 & 53.3 & 33.3 & 94.8 & 71.5 & 63.1 & 80.8 & 50.6 \\

% \Xhline{0.3 pt}
% \midrule
% \textbf{\textit{Re-implemented Models}} &   &  &  &  &  &  &  &  & & & & \\
% SenseNova-800k-Qwen2.5-VL-7B &  56.7  & 60.9 & 51 & 42 & 55.1 & 49.7 & 36.7 & 89.5 & 68.6 & 59.9 & & 47.1 \\
% Cambrains-590k-Qwen2.5-VL-7B &  55.1 & 60.3 & 48.1 & 38.2 & 55 & 50.4 & 29.5 & 89.6 & 69.3 & 60.6 &  & 50.0 \\
% VST-500k-Qwen2.5-VL-7B & 57.3 & 61.4 & 50.7 & 42.2 & 54.1 & 54.1 & 32.4 & 93.2 & 70.6 & 61.7 &  & 50.7 \\
% \Xhline{0.3 pt}
\midrule
\textbf{\textit{Ours}} &   &  &  &  &  &  &  &  &  & & & \\
OpenSpatial-InternVL2.5-8B &  59.3 (\textcolor{green!60!black}{+7.7}) & 63.5(\textcolor{green!60!black}{+8.6}) & 58.3(\textcolor{green!60!black}{+9.4}) & 43.0(\textcolor{green!60!black}{+2.2}) & 56.7(\textcolor{green!60!black}{+17.6}) & 52.0(\textcolor{green!60!black}{+1.0}) &  38.7(\textcolor{green!60!black}{+10.1}) & \cellcolor{blueLv3}93.8(\textcolor{green!60!black}{+13.9}) & \cellcolor{blueLv3}68.6(\textcolor{red!80!black}{-0.8}) & 57.4 & 78.5 & 48.0  \\
OpenSpatial-InternVL3-8B & \cellcolor{blueLv3}59.8(\textcolor{green!60!black}{+6.6})  & \cellcolor{blueLv3}66.0(\textcolor{green!60!black}{+10.3}) & 58.3 (\textcolor{green!60!black}{+7.8}) & \cellcolor{blueLv4}44.5(\textcolor{green!60!black}{+4.0}) & \cellcolor{blueLv3}57.4(\textcolor{green!60!black}{+18.7}) & \cellcolor{blueLv3}53.5(\textcolor{green!60!black}{+0.8}) & 38.6(\textcolor{green!60!black}{+7.7}) & 93.7(\textcolor{green!60!black}{+7.7}) & 66.6(\textcolor{red!80!black}{-3.9}) & 62.4 & 82.0 & 51.2 \\
OpenSpatial-Qwen2.5-VL-7B & 59.5(\textcolor{green!60!black}{+9.5})  & 65.9(\textcolor{green!60!black}{+10.6}) & \cellcolor{blueLv3}58.4(\textcolor{green!60!black}{+8.3})  & 41.8(\textcolor{green!60!black}{+0.8})  & 56.7 (\textcolor{green!60!black}{+20.7})  & 53.2(\textcolor{green!60!black}{+4.2}) & \cellcolor{blueLv3}39.6(\textcolor{green!60!black}{+13.1}) & 92.5(\textcolor{green!60!black}{+18.4}) & 68.3(\textcolor{green!60!black}{+0.2}) & 62.2 & 80.9 & 49.4 \\
OpenSpatial-Qwen3-VL-8B & \cellcolor{blueLv4}62.1(\textcolor{green!60!black}{+5.4})  & \cellcolor{blueLv4}68.2(\textcolor{green!60!black}{+2.1}) & \cellcolor{blueLv4}59.8(\textcolor{green!60!black}{+10.3}) &\cellcolor{blueLv3} 44.2(\textcolor{green!60!black}{+4.1}) & \cellcolor{blueLv4}61.6(\textcolor{green!60!black}{+6.0}) & \cellcolor{blueLv4}56.2(\textcolor{green!60!black}{+3.4}) & \cellcolor{blueLv4}41.9(\textcolor{green!60!black}{+13.8}) & \cellcolor{blueLv4}94.0(\textcolor{green!60!black}{+3.2}) & \cellcolor{blueLv4}71.0(\textcolor{green!60!black}{+0.3}) & 63.7 & 82.1 & 56.8 \\
% \Xhline{0.5 pt}
\bottomrule
\end{tabular}}
% \vspace{-0.5cm}
\end{table*}

\subsection{Quality Evaluation}
\subsubsection{Main Results:} As illustrated in Tab.~\ref{tab:main-results} , the OpenSpatial model series not only sets a new state-of-the-art on specialized spatial benchmarks but also preserves its versatility on general-purpose benchmarks without catastrophic forgetting. Upon integrating the data synthesized by our engine, we observe a substantial performance surge across all spatial reasoning tasks compared to the baseline, with improvements typically ranging from \textbf{5.4 to 9.5 points}. These consistent gains across diverse metrics suggest that OpenSpatial fosters a holistic understanding of 3D geometry rather than merely overfitting to specific spatial patterns. Particularly noteworthy are the results on BLINK, AllAngles, and MMSI, where our models achieve remarkable leaps \textbf{exceeding 10 points}. This significant margin allows us to outstrip existing spatial intelligence models by a wide gap, \textbf{serving as a strong testament to the unparalleled quality and diversity of our generated data}. From a model-centric perspective, we observe that Qwen3-VL-8B exhibits superior compatibility with our data. This can be attributed to the integration of SigLIP~\cite{siglip} as the vision encoder, which significantly bolsters the model’s visual perception and allows for a more nuanced spatial understanding of complex 3D scenes. However, we also identify a performance bottleneck in certain desktop-level and outdoor scenarios. This marginal improvement is primarily attributed to the current data distribution skew, and we intend to broaden our data coverage to encompass these complex environments in future iterations.

\begin{table*}[t]
\centering
\caption{Comparison with open-source datasets for spatial reasoning. `$-$' indicates the difference from the \textcolor{blueLv5}{best result}. MAD: Mean Deviation; Std. Dev.: Standard Deviation. \textbf{Note}: The subset of OpenSpatial is constructed independently and does not include any data from SenseNova-SI.}
\label{tab:data-compare}
% \vspace{-0.3cm}
% \tiny
\renewcommand{\arraystretch}{1} % 默认是1.0，小于1会变紧凑
\resizebox{\linewidth}{!}{
\begin{tabular}{l |c|c c|l  l l l l l l l l }
% \Xhline{0.5 pt}
\toprule
\textbf{Data source}& \textbf{Data size} & \textbf{MAD} & \textbf{Std. Dev.}   & \textbf{BLINK} & \textbf{AllAngles} & \textbf{ERQA} & \textbf{VSI} & \textbf{3DSR} & \textbf{MMSI} & \textbf{CV-3D} & \textbf{RealWorldQA} \\
% \Xhline{0.3 pt}
\midrule

Cambrain-S~\cite{cambrian-s} & 590k  & -6.0 & 5.4 & 54.1(\textcolor{red!80!black}{-10.1})  & 48.6 (\textcolor{red!80!black}{-5.3}) & 39.2(\textcolor{red!80!black}{-3.0}) & \cellcolor{blueLv4}57.0 & 49.7(\textcolor{red!80!black}{-2.3}) & 29.2(\textcolor{red!80!black}{-7.1}) & 75.3(\textcolor{red!80!black}{-17.9}) & 68.1(\textcolor{red!80!black}{-2.6}) \\
SenseNova-SI~\cite{sensenova-si} & 800k  & -6.5 & 7.0 & 59.7 (\textcolor{red!80!black}{-4.5})  & 47.5(\textcolor{red!80!black}{-6.4}) & 38.0(\textcolor{red!80!black}{-4.2}) & 55.0(\textcolor{red!80!black}{-2.0}) & 48.9 (\textcolor{red!80!black}{-5.1}) & \cellcolor{blueLv4}36.3 & 69.0(\textcolor{red!80!black}{-24.2}) & 65.1(\textcolor{red!80!black}{-5.6}) \\
VST~\cite{VST} & 500k  & -2.8 & 3.9 &  61.4(\textcolor{red!80!black}{-2.8}) & 50.7(\textcolor{red!80!black}{-3.2}) & \cellcolor{blueLv4}42.2 & 44.6(\textcolor{red!80!black}{-12.4}) & \cellcolor{blueLv4}54.1 & 32.4(\textcolor{red!80!black}{-3.9}) & \cellcolor{blueLv4}93.2 & 70.6(\textcolor{red!80!black}{-0.1}) \\
OpenSpatial(subset) & 500k  & -2.5 & 4.4 & \cellcolor{blueLv4}64.2 & \cellcolor{blueLv4}53.9 & 41.3(\textcolor{red!80!black}{-0.9}) & 43.0(\textcolor{red!80!black}{-14.0}) & 52.0(\textcolor{red!80!black}{-2.1}) & 34.7(\textcolor{red!80!black}{-1.6}) & 91.8(\textcolor{red!80!black}{-1.4}) & \cellcolor{blueLv4}70.7 \\

\bottomrule
\end{tabular}}

\vspace{-0.3cm}
\end{table*}

% \vspace{1cm}
\subsubsection{Comparative Study with Open-source Data:} For a fair quality assessment, we benchmarked a scale-matched subset of OpenSpatial against open-source datasets. We specifically excluded SenseNova-SI-800k to focus on the intrinsic quality of our generated data. Following a unified training protocol based on Qwen2.5-VL, the comparative results are reported in Tab.~\ref{tab:data-compare}. Statistical analysis reveals that OpenSpatial and VST both emerge as versatile and well-rounded datasets, achieving competitive stability and the narrowest mean gaps ($-2.5$ and $-2.8$, respectively) across all benchmarks. In contrast, while Cambrian-S and SenseNova-SI show larger overall fluctuations (Mean Deviation: $-6.0$ to $-6.5$ and Standard Deviation: $5.4$ to $7.0$), they exhibit specialized proficiency in niche domains, such as the VSI and MMSI benchmarks. This performance pattern highlights a clear data complementarity: while our engine prioritizes a broad and consistent spatial understanding. SenseNova-SI-800k exhibits localized strengths on specific benchmarks that complement our broad coverage, thus we integrate it in OpenSpatial to further bolster performance in specialized scenarios and achieve a more versatile spatial understanding

% \vspace{-0.6cm}
% \begin{table*}[h]
% % 1. 调整比例：左侧加宽到 0.65，右侧收窄到 0.33
% \caption{Rationality Analysis of Module Design.}
% \label{tab:module}
% \vspace{-0.3cm}
% \begin{minipage}{.6\textwidth}
% \centering
% \resizebox{\textwidth}{!}{ 
%     \tablestyle{8pt}{1.1} % 列数少了，可以稍微加大间距
%     \begin{tabular}{l | c c c c  }
%     \toprule
%     \textbf{Setting: data size(200k)} & \textbf{BLINK} & \textbf{AllAngles} & \textbf{VSI} & \textbf{CV-3D}  \\
%     \midrule
%     Qwen2.5-VL-7B  & 55.3 & 50.1 & 36.0 &  73.8 \\
%     +Point Cloud Centric  & 57.2 & 49.7 & 37.2 &  83.7 \\
%     +3D-Box Centric & 60.3 & 53.2 & 41.7 &  89.9 \\
%     +3D-Box Centric (no filter) & 56.6 & 47.0 & 32.1 &  78.2 \\
%     \bottomrule
%     \end{tabular}
% }
% \end{minipage}\hfill
% % 2. 右侧 minipage 变窄
% \begin{minipage}{0.5\textwidth}
% % \centering
% % \vspace{.9em}
% % 3. 进一步限制图片宽度，比如 0.9 倍当前 minipage 宽度，或指定具体高度
% \includegraphics[width=0.7\linewidth]{Fig/module_resonable.png} 
% \end{minipage}
% % \vspace{-.5em}

% \vspace{-1.0cm}
% \end{table*}
\vspace{-0.3cm}
\subsubsection{Module Reasonability:} In this section, we evaluate the effectiveness of individual modules within our data engine. To ensure consistency, we reproduce the ablation datasets using ScanNet as the source, with each experimental set maintained at approximately 200k samples. As shown in Tab.~\ref{tab:module}, shifting from a box-centric to a point-cloud-centric representation leads to varying degrees of performance degradation. This is further illustrated by the qualitative examples on the right side of the table: partial point clouds fail to represent the complete geometry of objects, leading to inaccurate data generation, particularly for spatial measurement tasks. Furthermore, the filtering mechanism is crucial for box-centric design. Due to visual occlusion (exemplified by the red boxes), failing to filter out such cases introduces hallucinations and significantly undermines model performance.

\vspace{-0.6cm}
\begin{table*}[t]
% 1. 调整比例：左侧加宽到 0.65，右侧收窄到 0.33
\caption{Rationality Analysis of Module Design.}
\label{tab:module}
\vspace{-0.3cm}
\begin{minipage}{.6\textwidth}
\centering
\resizebox{\textwidth}{!}{ 
    \tablestyle{8pt}{1.1} % 列数少了，可以稍微加大间距
    \begin{tabular}{l | c c c c  }
    \toprule
    \textbf{Setting: data size(200k)} & \textbf{BLINK} & \textbf{AllAngles} & \textbf{VSI} & \textbf{CV-3D}  \\
    \midrule
    Qwen2.5-VL-7B  & 55.3 & 50.1 & 36.0 &  73.8 \\
    +Point Cloud Centric  & 57.2 & 49.7 & 37.2 &  83.7 \\
    +3D-Box Centric & 60.3 & 53.2 & 41.7 &  89.9 \\
    +3D-Box Centric (no filter) & 56.6 & 47.0 & 32.1 &  78.2 \\
    \bottomrule
    \end{tabular}
}
\end{minipage}\hfill
% 2. 右侧 minipage 变窄
\begin{minipage}{0.5\textwidth}
% \centering
% \vspace{.9em}
% 3. 进一步限制图片宽度，比如 0.9 倍当前 minipage 宽度，或指定具体高度
\includegraphics[width=0.7\linewidth]{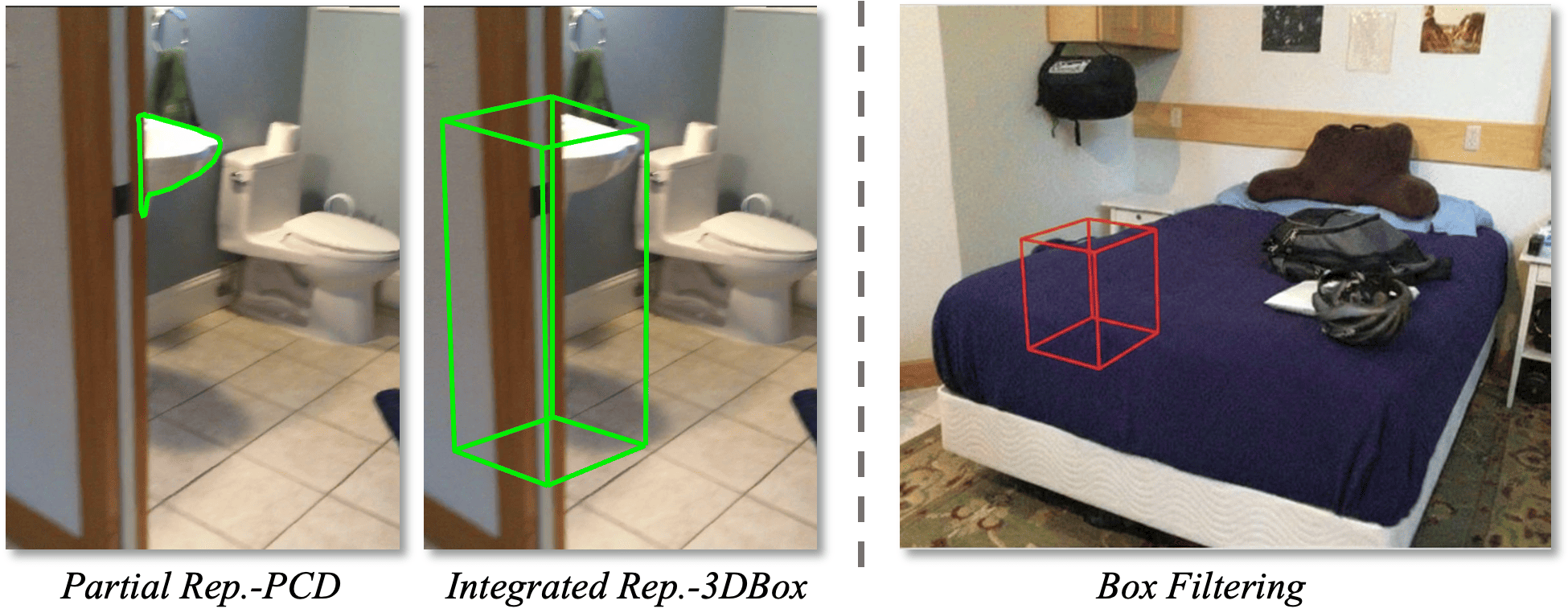} 
\end{minipage}
% \vspace{-.5em}

% \vspace{-0.5cm}
\end{table*}

% \vspace{0.3cm}

\begin{table*}[h]
\centering
\caption{Ablation on data scaling. \textbf{Color Map}: \colorbox{blueLv1}{worst} to \colorbox{blueLv5}{best}.}
\label{tab:data-scale}
\vspace{-0.3cm}
% \tiny
\renewcommand{\arraystretch}{1} % 默认是1.0，小于1会变紧凑
\resizebox{\linewidth}{!}{
\tablestyle{8pt}{1.0}
\begin{tabular}{l | c | c c c c c c c c }
% \Xhline{0.5 pt}
\toprule
\textbf{Data Size} & \textbf{3D-Avg} & \textbf{BLINK} & \textbf{AllAngles} & \textbf{ERQA} & \textbf{VSI} & \textbf{3DSR} & \textbf{MMSI} & \textbf{CV-3D} & \textbf{RealWorldQA} \\
% \Xhline{0.3 pt}
\midrule

20\% & \cellcolor{blueLv1}  57.6 &  64.9  & 55.0  & 41.0  & 51.8  & 52.0 &  34.7  & 91.8  & 69.5  \\
40\% & \cellcolor{blueLv2}  58.5 &  66.2  &  56.8 & 42.1  &  52.4 &  52.9 &  35.8 & 91.0  & 71.1 \\
60\% & \cellcolor{blueLv3}  58.6  &  66.9  & 56.7  &  42.0  & 53.2  & 52.3  & 36.0  &  92.1 & 70.7 \\
80\% & \cellcolor{blueLv4} 59.2  &  66.2  & 59.7  & 41.2  & 53.5  & 53.0  & 37.6  & 91.5  & 70.5 \\
Full & \cellcolor{blueLv5} 59.7 &  65.9  &  58.4 &  41.8 &  56.7 &  54.3 &  39.6 &  92.5 &  68.3 \\

\bottomrule
\end{tabular}}

% \vspace{-1.0cm}
\end{table*}

\begin{table*}[h]
\centering
\caption{Ablation on model scaling. \colorbox{blueLv4}{Dark blue}: best result. \colorbox{blueLv3}{Light blue}: second best.}
\label{tab:model-scale}
\vspace{-0.3cm}
% \tiny
\renewcommand{\arraystretch}{1} % 默认是1.0，小于1会变紧凑
\resizebox{\linewidth}{!}{
\tablestyle{5pt}{1.0}
\begin{tabular}{l | c | c c c c c c c c }
% \Xhline{0.5 pt}
\toprule
\textbf{Model Size} & \textbf{3D-Avg} & \textbf{BLINK} & \textbf{AllAngles} & \textbf{ERQA} & \textbf{VSI} & \textbf{3DSR} & \textbf{MMSI} & \textbf{CV-3D} & \textbf{RealWorldQA} \\
% \Xhline{0.3 pt}
\midrule
OpenSpatial-Qwen2.5-VL-3B  & 56.1 & 61.0 & 53.0 & 40.0 & 55.2 & 47.8 & 32.0 & \cellcolor{blueLv4}94.0 & 66.0 \\
OpenSpatial-Qwen2.5-VL-7B  & \cellcolor{blueLv3}59.7 & \cellcolor{blueLv3}65.9 & \cellcolor{blueLv3}58.4 & \cellcolor{blueLv3}41.8 & \cellcolor{blueLv3}56.7 & \cellcolor{blueLv3}54.3 & \cellcolor{blueLv3}39.6 & \cellcolor{blueLv3}92.5 & \cellcolor{blueLv3}68.3 \\
OpenSpatial-Qwen2.5-VL-32B  & \cellcolor{blueLv4}61.3 & \cellcolor{blueLv4}68.2 & \cellcolor{blueLv4}63.3 & \cellcolor{blueLv4}44.0 & \cellcolor{blueLv4}57.3 & \cellcolor{blueLv4}55.3 & \cellcolor{blueLv4}39.8 & 93.4 & \cellcolor{blueLv4}69.3 \\

\bottomrule
\end{tabular}}
% \vspace{-1cm}
\end{table*}

\subsection{Scalability Evaluation}
\subsubsection{Data Scaling:} We perform category-wise downsampling on both spatial reasoning data and general data, utilizing Qwen2.5-VL as our base model. The results, as summarized in Tab.~\ref{tab:data-scale}, reveal that while individual benchmarks may not exhibit a strictly monotonic increase in performance with data scaling, the 3D-Avg metric shows a consistent positive correlation. This trend suggests that \textbf{increasing data volume systematically enhances the model’s comprehensive spatial reasoning capabilities}. However, we also observe that the rate of performance gain diminishes as the data scale grows, indicating that \textbf{further improvements in spatial intelligence require exponentially larger datasets}.

\begin{figure}[h]
\begin{center}
   \includegraphics[width=1.0\linewidth]{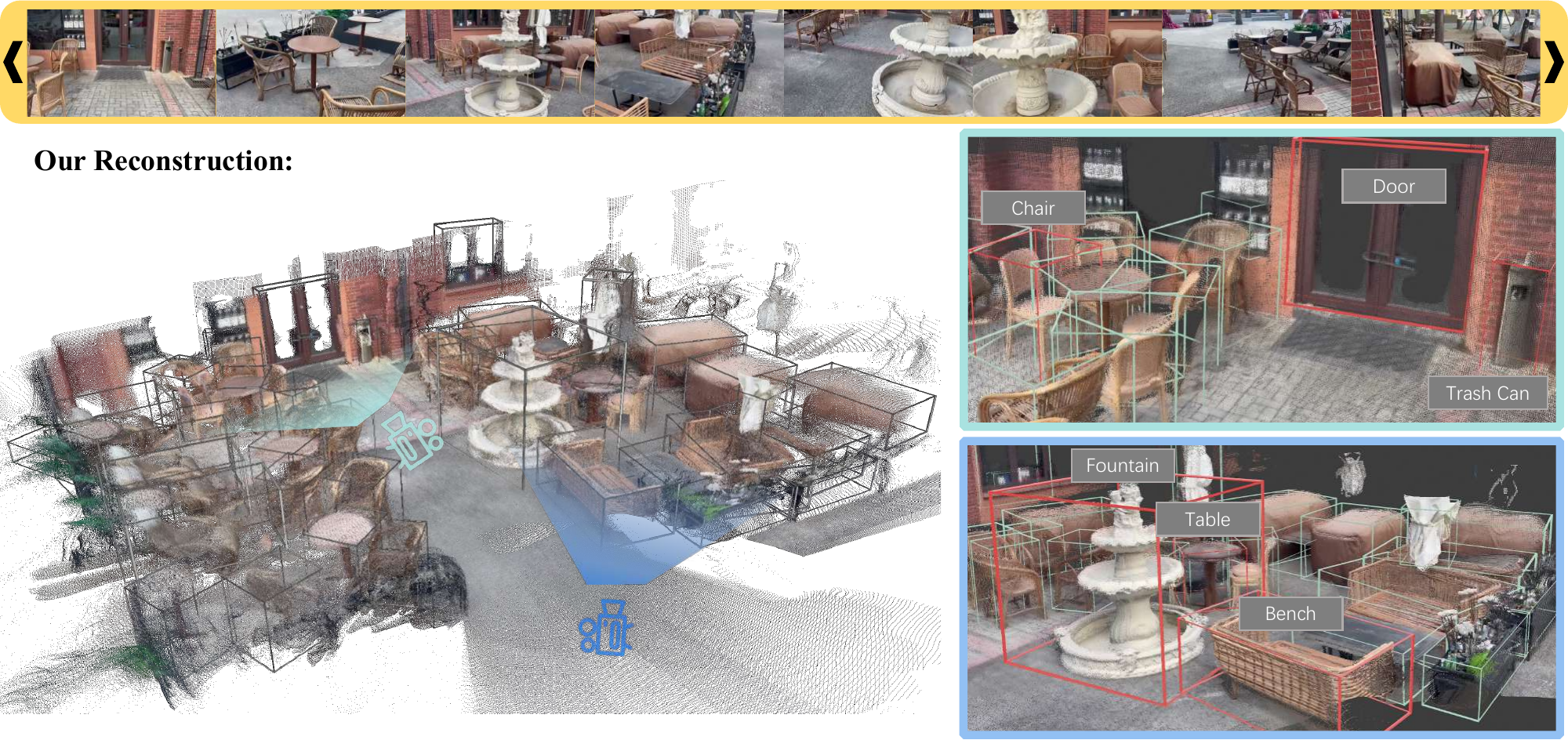}
   \caption{Visualization of 3D lifting results from in-the-wild outdoor web data.} 
   \label{fig:3d-lifting}
\end{center}
\vspace{-1cm}
\end{figure}

\subsubsection{Data Source Scaling:} Existing 3D datasets are constrained by a limited number of scenes and a heavy bias towards indoor environments. To further scale the data volume and enhance scene diversity, we develop a robust 3D lifting pipeline capable of reconstructing 3D scenes from in-the-wild video data, while simultaneously generating comprehensive annotations including semantic tags, masks, and 3D bounding boxes. Fig.~\ref{fig:3d-lifting} illustrates our annotation results on an uncurated outdoor video. As shown, our pipeline not only recovers the scene geometry (point clouds) with high fidelity but also produces accurate tags and boxes, effectively meeting the stringent requirements for 3D spatial understanding data production. Leveraging this pipeline, we are able to significantly scale our dataset to unprecedented diversity and volume. Tab.~\ref{tab:3d_lifting} demonstrates the effectiveness of the spatial understanding data produced solely from web-sourced data via our 3D lifting pipeline.

\subsubsection{Model Scaling:} Investigating model size is equally crucial for spatial understanding, as models with larger parameter scales possess greater representative capacity to internalize and organize complex spatial knowledge. 
As illustrated in Tab.~\ref{tab:data-scale}, we evaluate the performance of Qwen2.5-VL across various scales, including 3B, 7B, and 32B parameters, under identical data configurations. It is evident that nearly all evaluation metrics exhibit a consistent upward trend as the model size increases, highlighting the positive impact of model capacity on spatial reasoning tasks. These results further validate the significance of a scalable data engine. By consistently translating increased data volume into performance gains, our engine demonstrates its ability to provide the high-quality supervision necessary for advancing the spatial intelligence of VLMs.

\begin{figure}[htbp]
\begin{center}
   \includegraphics[width=1.0\linewidth]{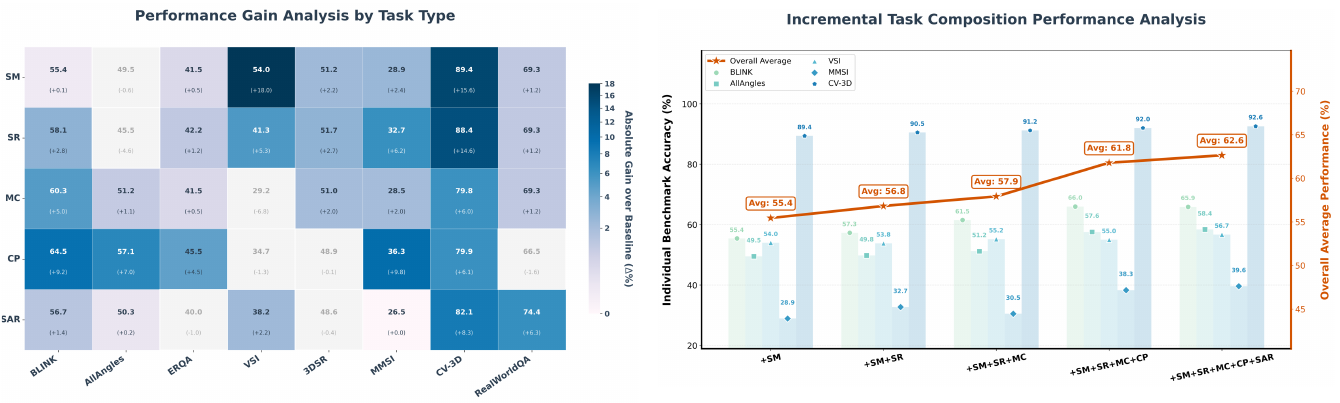}
   \caption{Impact of diverse tasks on spatial intelligence. Best zoomed in.} 
   \label{fig:task-diverse}
\end{center}
\vspace{-1.0cm}
\end{figure}

\subsection{Diversity Evaluation}
To provide a granular understanding of how task diversity steers the evolution of spatial reasoning, we conducted a dual-faceted evaluation as illustrated in Fig.~\ref{fig:task-diverse}. 

\noindent \textbf{Task-Specific Contributions and Complementarity}: As depicted in the left heatmap of Fig.~\ref{fig:task-diverse}, individual tasks exhibit heterogeneous performance footprints across diverse benchmarks. For instance, tasks focused on Spatial Measurement (SM) yield substantial gains in metric-heavy evaluations, whereas Camera Perception (CP) tasks predominantly bolster the model’s ability to decode extrinsic parameters and ego-motion, leading to significant improvements in benchmarks requiring precise viewpoint awareness. This divergence underscores that our task library is not redundant but rather possesses a strong complementary architecture, where each task targets a unique dimension of spatial cognition. 

\noindent \textbf{Incremental Synergy and Compositional Trends}: Moving to the right panel of Fig.~\ref{fig:task-diverse}, we investigate the cumulative impact of incremental task integration. The results reveal a compelling compositional synergy: with the increment of the task diversity, the model's comprehensive capabilities scale accordingly. We observe occasional, localized performance plateaus or slight "dips" upon the introduction of certain task combinations; these are likely attributed to data distribution shifts or the interference of gradient directions during multi-task optimization. However, the overarching trajectory remains unmistakably positive. The orange "Overall Average" curve maintains a steady and robust ascent, signifying that increased task diversity effectively mitigates the limitations of single-task learning and fosters a more holistic and resilient spatial intelligence.

% \subsubsection{Task diversity}

% \subsubsection{QA diversity}

% \begin{figure}[h]
% \begin{center}
%    \includegraphics[width=0.5\linewidth]{Fig/efficiency.pdf}
%    \caption{Efficiency breakdown of our data production pipeline.} 
%    \label{fig:efficiency}
% \end{center}
% \vspace{-1cm}
% \end{figure}

\vspace{-0.7cm}
\begin{figure}[htbp]
  \centering
  
  % 左侧：表格部分
  \begin{minipage}[h]{0.52\textwidth} % 调整宽度比例
    \centering
    \captionof{table}{Effectiveness of data source scaling. We independently validate the effectiveness of our 3D lifting data.} % 表格标题
    \label{tab:3d_lifting}
    \vspace{0.2cm}
    % \small % 如果表格太宽，可以缩小字号
    \scalebox{0.75}{
    \begin{tabular}{l | c c c c c}
      \toprule
      % \textbf{Setting: 200k} & \textbf{BLINK} & \textbf{AllAngles} & \textbf{3DSR} & \textbf{CV-3D} & \textbf{RealWorldQA} \\
      % \midrule
      % Qwen2.5-VL-7B               & 55.3 & 50.1 & 49.0 & 73.8 & 68.1 \\
      % +3D Lifting        & 60.0 & 52.2 & 51.5 & 87.3 & 72.2 \\
      \textbf{Setting: 200k} & \textbf{BLINK} & \textbf{3DSR} & \textbf{CV-3D} & \textbf{RealWorldQA} \\
      \midrule
      Qwen2.5-VL-7B               & 55.3 & 49.0 & 73.8 & 68.1 \\
      +3D Lifting        & 62.2 & 54.3 & 87.9 & 71.8 \\
      \bottomrule
    \end{tabular}
    }

  \end{minipage}
  % \hfill % 撑开中间间距
  % 右侧：图片部分
  \begin{minipage}[h]{0.45\textwidth} % 调整宽度比例
    \centering
    % \vspace{0.5cm}
    \includegraphics[width=1\linewidth]{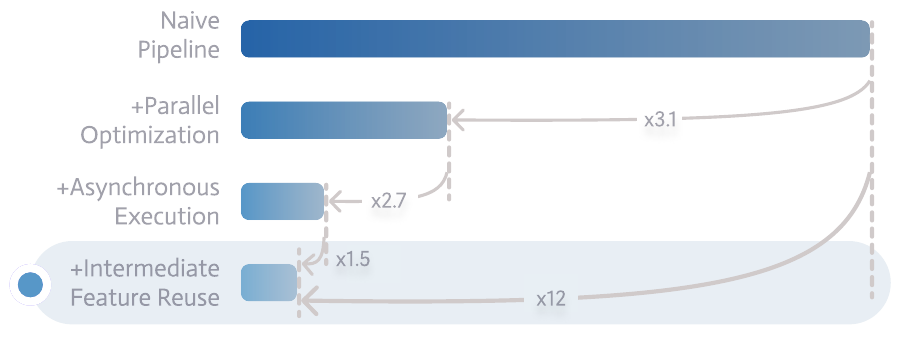}
    \vspace{-17pt}
    \captionof{figure}{Efficiency breakdown.} % 图片标题
    \vspace{-0.3cm}
    \label{fig:efficiency}
    % \vspace{10pt}
  \end{minipage}
\vspace{-0.7cm}
\end{figure}

\subsection{Efficiency Evaluation}

To enhance the efficiency of our data production pipeline, we implemented a series of systematic optimizations. First, parallel processing was applied across most components to maximize throughput. Second, we leveraged message queues to enable asynchronous execution between consecutive stages; this pipelining strategy allows the current stage to perform inference on a batch while the preceding stage simultaneously processes the next. Finally, for tasks sharing common intermediate features, we developed an automatic reuse mechanism to avoid redundant computations and further streamline the workflow. The performance gains resulting from these efficiency optimizations are illustrated in Fig.~\ref{fig:efficiency}.

\section{Conclusion}
In this work, we introduce OpenSpatial, a principled data engine that shifts the focus from static datasets to a transparent, scalable infrastructure for spatial intelligence. By establishing a 3D box-centric paradigm, the OpenSpatial engine serves as a vital bridge between sparse 2D visual cues and intrinsic 3D metric properties, providing a viewpoint-invariant foundation that was previously confined to closed-source pipelines. This engine not only enables the synthesis of our OpenSpatial-3M dataset—which achieves state-of-the-art performance across diverse MLLM architectures—but more importantly, it establishes a sustainable foundation for producing diverse spatial data across multiple sources organized into five hierarchical task categories, allowing for continuous expansion and refinement of spatial understanding. By open-sourcing OpenSpatial, we aim to democratize the creation of high-quality data, serving as a robust cornerstone for the community to advance embodied AI and robotics.

% \section*{Acknowledgements}

\bibliographystyle{splncs04}
\bibliography{main}

@String(CVPR  = {IEEE Conf. Comput. Vis. Pattern Recog.})

@String(ICCV  = {Int. Conf. Comput. Vis.})

@String(CVPR  = {CVPR})

@String(ICCV  = {ICCV})

@inproceedings{SAM,
  title={Segment anything},
  author={Kirillov, Alexander and Mintun, Eric and Ravi, Nikhila and Mao, Hanzi and Rolland, Chloe and Gustafson, Laura and Xiao, Tete and Whitehead, Spencer and Berg, Alexander C and Lo, Wan-Yen and others},
  booktitle={Proceedings of the IEEE/CVF international conference on computer vision},
  pages={4015--4026},
  year={2023}
}

@article{VST,
  title={Visual spatial tuning},
  author={Yang, Rui and Zhu, Ziyu and Li, Yanwei and Huang, Jingjia and Yan, Shen and Zhou, Siyuan and Liu, Zhe and Li, Xiangtai and Li, Shuangye and Wang, Wenqian and others},
  journal={arXiv preprint arXiv:2511.05491},
  year={2025}
}

@inproceedings{embodiedscan,
  title={Embodiedscan: A holistic multi-modal 3d perception suite towards embodied ai},
  author={Wang, Tai and Mao, Xiaohan and Zhu, Chenming and Xu, Runsen and Lyu, Ruiyuan and Li, Peisen and Chen, Xiao and Zhang, Wenwei and Chen, Kai and Xue, Tianfan and others},
  booktitle={Proceedings of the IEEE/CVF Conference on Computer Vision and Pattern Recognition},
  pages={19757--19767},
  year={2024}
}

@inproceedings{scannet,
  title={Scannet: Richly-annotated 3d reconstructions of indoor scenes},
  author={Dai, Angela and Chang, Angel X and Savva, Manolis and Halber, Maciej and Funkhouser, Thomas and Nie{\ss}ner, Matthias},
  booktitle={Proceedings of the IEEE conference on computer vision and pattern recognition},
  pages={5828--5839},
  year={2017}
}

@inproceedings{scannet++,
  title={Scannet++: A high-fidelity dataset of 3d indoor scenes},
  author={Yeshwanth, Chandan and Liu, Yueh-Cheng and Nie{\ss}ner, Matthias and Dai, Angela},
  booktitle={Proceedings of the IEEE/CVF International Conference on Computer Vision},
  pages={12--22},
  year={2023}
}

@article{matterport3d,
  title={Matterport3d: Learning from rgb-d data in indoor environments},
  author={Chang, Angel and Dai, Angela and Funkhouser, Thomas and Halber, Maciej and Niessner, Matthias and Savva, Manolis and Song, Shuran and Zeng, Andy and Zhang, Yinda},
  journal={arXiv preprint arXiv:1709.06158},
  year={2017}
}

@inproceedings{sunrgbd,
  title={Sun rgb-d: A rgb-d scene understanding benchmark suite},
  author={Song, Shuran and Lichtenberg, Samuel P and Xiao, Jianxiong},
  booktitle={Proceedings of the IEEE conference on computer vision and pattern recognition},
  pages={567--576},
  year={2015}
}

@article{arkitscenes,
  title={Arkitscenes: A diverse real-world dataset for 3d indoor scene understanding using mobile rgb-d data},
  author={Baruch, Gilad and Chen, Zhuoyuan and Dehghan, Afshin and Dimry, Tal and Feigin, Yuri and Fu, Peter and Gebauer, Thomas and Joffe, Brandon and Kurz, Daniel and Schwartz, Arik and others},
  journal={arXiv preprint arXiv:2111.08897},
  year={2021}
}

@inproceedings{hypersim,
  title={Hypersim: A photorealistic synthetic dataset for holistic indoor scene understanding},
  author={Roberts, Mike and Ramapuram, Jason and Ranjan, Anurag and Kumar, Atulit and Bautista, Miguel Angel and Paczan, Nathan and Webb, Russ and Susskind, Joshua M},
  booktitle={Proceedings of the IEEE/CVF international conference on computer vision},
  pages={10912--10922},
  year={2021}
}

@article{llava,
  title={Visual instruction tuning},
  author={Liu, Haotian and Li, Chunyuan and Wu, Qingyang and Lee, Yong Jae},
  journal={Advances in neural information processing systems},
  volume={36},
  pages={34892--34916},
  year={2023}
}

@article{qwen2,
  title={Qwen2-vl: Enhancing vision-language model's perception of the world at any resolution},
  author={Wang, Peng and Bai, Shuai and Tan, Sinan and Wang, Shijie and Fan, Zhihao and Bai, Jinze and Chen, Keqin and Liu, Xuejing and Wang, Jialin and Ge, Wenbin and others},
  journal={arXiv preprint arXiv:2409.12191},
  year={2024}
}

@article{qwen3,
  title={Qwen3-VL-Embedding and Qwen3-VL-Reranker: A Unified Framework for State-of-the-Art Multimodal Retrieval and Ranking},
  author={Li, Mingxin and Zhang, Yanzhao and Long, Dingkun and Chen, Keqin and Song, Sibo and Bai, Shuai and Yang, Zhibo and Xie, Pengjun and Yang, An and Liu, Dayiheng and others},
  journal={arXiv preprint arXiv:2601.04720},
  year={2026}
}

@article{qwen3-t,
  title={Qwen3-vl technical report},
  author={Bai, Shuai and Cai, Yuxuan and Chen, Ruizhe and Chen, Keqin and Chen, Xionghui and Cheng, Zesen and Deng, Lianghao and Ding, Wei and Gao, Chang and Ge, Chunjiang and others},
  journal={arXiv preprint arXiv:2511.21631},
  year={2025}
}

@article{qwen,
  title={Qwen-vl: A versatile vision-language model for understanding, localization},
  author={Bai, Jinze and Bai, Shuai and Yang, Shusheng and Wang, Shijie and Tan, Sinan and Wang, Peng and Lin, Junyang and Zhou, Chang and Zhou, Jingren},
  journal={Text Reading, and Beyond},
  volume={2},
  number={1},
  pages={1},
  year={2023}
}

@inproceedings{internvl,
  title={Internvl: Scaling up vision foundation models and aligning for generic visual-linguistic tasks},
  author={Chen, Zhe and Wu, Jiannan and Wang, Wenhai and Su, Weijie and Chen, Guo and Xing, Sen and Zhong, Muyan and Zhang, Qinglong and Zhu, Xizhou and Lu, Lewei and others},
  booktitle={Proceedings of the IEEE/CVF conference on computer vision and pattern recognition},
  pages={24185--24198},
  year={2024}
}

@article{internvl3.5,
  title={Internvl3. 5: Advancing open-source multimodal models in versatility, reasoning, and efficiency},
  author={Wang, Weiyun and Gao, Zhangwei and Gu, Lixin and Pu, Hengjun and Cui, Long and Wei, Xingguang and Liu, Zhaoyang and Jing, Linglin and Ye, Shenglong and Shao, Jie and others},
  journal={arXiv preprint arXiv:2508.18265},
  year={2025}
}

@article{internvl3,
  title={Internvl3: Exploring advanced training and test-time recipes for open-source multimodal models},
  author={Zhu, Jinguo and Wang, Weiyun and Chen, Zhe and Liu, Zhaoyang and Ye, Shenglong and Gu, Lixin and Tian, Hao and Duan, Yuchen and Su, Weijie and Shao, Jie and others},
  journal={arXiv preprint arXiv:2504.10479},
  year={2025}
}

@article{instructblip,
  title={Instructblip: Towards general-purpose vision-language models with instruction tuning},
  author={Dai, Wenliang and Li, Junnan and Li, Dongxu and Tiong, Anthony and Zhao, Junqi and Wang, Weisheng and Li, Boyang and Fung, Pascale N and Hoi, Steven},
  journal={Advances in neural information processing systems},
  volume={36},
  pages={49250--49267},
  year={2023}
}

@article{spatialmllm,
  title={Spatial-mllm: Boosting mllm capabilities in visual-based spatial intelligence},
  author={Wu, Diankun and Liu, Fangfu and Hung, Yi-Hsin and Duan, Yueqi},
  journal={arXiv preprint arXiv:2505.23747},
  year={2025}
}

@article{vlm-3r,
  title={Vlm-3r: Vision-language models augmented with instruction-aligned 3d reconstruction},
  author={Fan, Zhiwen and Zhang, Jian and Li, Renjie and Zhang, Junge and Chen, Runjin and Hu, Hezhen and Wang, Kevin and Qu, Huaizhi and Wang, Dilin and Yan, Zhicheng and others},
  journal={arXiv preprint arXiv:2505.20279},
  year={2025}
}

@article{3Dthink,
  title={Think with 3d: Geometric imagination grounded spatial reasoning from limited views},
  author={Chen, Zhangquan and Zhang, Manyuan and Yu, Xinlei and Luo, Xufang and Sun, Mingze and Pan, Zihao and Feng, Yan and Pei, Peng and Cai, Xunliang and Huang, Ruqi},
  journal={arXiv preprint arXiv:2510.18632},
  year={2025}
}

@inproceedings{spatialbot,
  title={Spatialbot: Precise spatial understanding with vision language models},
  author={Cai, Wenxiao and Ponomarenko, Iaroslav and Yuan, Jianhao and Li, Xiaoqi and Yang, Wankou and Dong, Hao and Zhao, Bo},
  booktitle={2025 IEEE International Conference on Robotics and Automation (ICRA)},
  pages={9490--9498},
  year={2025},
  organization={IEEE}
}

@article{wu2025reinforcing,
  title={Reinforcing spatial reasoning in vision-language models with interwoven thinking and visual drawing},
  author={Wu, Junfei and Guan, Jian and Feng, Kaituo and Liu, Qiang and Wu, Shu and Wang, Liang and Wu, Wei and Tan, Tieniu},
  journal={arXiv preprint arXiv:2506.09965},
  year={2025}
}

@InProceedings{spatial-vlm,
    author    = {Chen, Boyuan and Xu, Zhuo and Kirmani, Sean and Ichter, Brain and Sadigh, Dorsa and Guibas, Leonidas and Xia, Fei},
    title     = {SpatialVLM: Endowing Vision-Language Models with Spatial Reasoning Capabilities},
    booktitle = {Proceedings of the IEEE/CVF Conference on Computer Vision and Pattern Recognition (CVPR)},
    month     = {June},
    year      = {2024},
    pages     = {14455-14465}
}

@article{spatialrgpt,
  title={Spatialrgpt: Grounded spatial reasoning in vision-language models},
  author={Cheng, An-Chieh and Yin, Hongxu and Fu, Yang and Guo, Qiushan and Yang, Ruihan and Kautz, Jan and Wang, Xiaolong and Liu, Sifei},
  journal={Advances in Neural Information Processing Systems},
  volume={37},
  pages={135062--135093},
  year={2024}
}

@article{spacer,
  title={Spacer: Reinforcing mllms in video spatial reasoning},
  author={Ouyang, Kun and Liu, Yuanxin and Wu, Haoning and Liu, Yi and Zhou, Hao and Zhou, Jie and Meng, Fandong and Sun, Xu},
  journal={arXiv preprint arXiv:2504.01805},
  year={2025}
}

@article{spatialladder,
  title={Spatialladder: Progressive training for spatial reasoning in vision-language models},
  author={Li, Hongxing and Li, Dingming and Wang, Zixuan and Yan, Yuchen and Wu, Hang and Zhang, Wenqi and Shen, Yongliang and Lu, Weiming and Xiao, Jun and Zhuang, Yueting},
  journal={arXiv preprint arXiv:2510.08531},
  year={2025}
}

@article{cambrian-s,
  title={Cambrian-s: Towards spatial supersensing in video},
  author={Yang, Shusheng and Yang, Jihan and Huang, Pinzhi and Brown, Ellis and Yang, Zihao and Yu, Yue and Tong, Shengbang and Zheng, Zihan and Xu, Yifan and Wang, Muhan and others},
  journal={arXiv preprint arXiv:2511.04670},
  year={2025}
}

@inproceedings{mindcube,
  title={Spatial mental modeling from limited views},
  author={Yin, Baiqiao and Wang, Qineng and Zhang, Pingyue and Zhang, Jianshu and Wang, Kangrui and Wang, Zihan and Zhang, Jieyu and Chandrasegaran, Keshigeyan and Liu, Han and Krishna, Ranjay and others},
  booktitle={Structural Priors for Vision Workshop at ICCV'25},
  year={2025}
}

@article{sensenova-si,
  title={Scaling spatial intelligence with multimodal foundation models},
  author={Cai, Zhongang and Wang, Ruisi and Gu, Chenyang and Pu, Fanyi and Xu, Junxiang and Wang, Yubo and Yin, Wanqi and Yang, Zhitao and Wei, Chen and Sun, Qingping and others},
  journal={arXiv preprint arXiv:2511.13719},
  year={2025}
}

@article{llava-ov,
  title={Llava-onevision: Easy visual task transfer},
  author={Li, Bo and Zhang, Yuanhan and Guo, Dong and Zhang, Renrui and Li, Feng and Zhang, Hao and Zhang, Kaichen and Zhang, Peiyuan and Li, Yanwei and Liu, Ziwei and others},
  journal={arXiv preprint arXiv:2408.03326},
  year={2024}
}

@inproceedings{blink,
  title={Blink: Multimodal large language models can see but not perceive},
  author={Fu, Xingyu and Hu, Yushi and Li, Bangzheng and Feng, Yu and Wang, Haoyu and Lin, Xudong and Roth, Dan and Smith, Noah A and Ma, Wei-Chiu and Krishna, Ranjay},
  booktitle={European Conference on Computer Vision},
  pages={148--166},
  year={2024},
  organization={Springer}
}

@article{allangles,
  title={Seeing from another perspective: Evaluating multi-view understanding in mllms},
  author={Yeh, Chun-Hsiao and Wang, Chenyu and Tong, Shengbang and Cheng, Ta-Ying and Wang, Ruoyu and Chu, Tianzhe and Zhai, Yuexiang and Chen, Yubei and Gao, Shenghua and Ma, Yi},
  journal={arXiv preprint arXiv:2504.15280},
  year={2025}
}

@inproceedings{vsi,
  title={Thinking in space: How multimodal large language models see, remember, and recall spaces},
  author={Yang, Jihan and Yang, Shusheng and Gupta, Anjali W and Han, Rilyn and Fei-Fei, Li and Xie, Saining},
  booktitle={Proceedings of the Computer Vision and Pattern Recognition Conference},
  pages={10632--10643},
  year={2025}
}

@inproceedings{3dsrbench,
  title={3dsrbench: A comprehensive 3d spatial reasoning benchmark},
  author={Ma, Wufei and Chen, Haoyu and Zhang, Guofeng and Chou, Yu-Cheng and Chen, Jieneng and de Melo, Celso and Yuille, Alan},
  booktitle={Proceedings of the IEEE/CVF International Conference on Computer Vision},
  pages={6924--6934},
  year={2025}
}

@article{mmsi,
  title={Mmsi-bench: A benchmark for multi-image spatial intelligence},
  author={Yang, Sihan and Xu, Runsen and Xie, Yiman and Yang, Sizhe and Li, Mo and Lin, Jingli and Zhu, Chenming and Chen, Xiaochen and Duan, Haodong and Yue, Xiangyu and others},
  journal={arXiv preprint arXiv:2505.23764},
  year={2025}
}

@article{cv-bench,
  title={Cambrian-1: A fully open, vision-centric exploration of multimodal llms},
  author={Tong, Peter and Brown, Ellis and Wu, Penghao and Woo, Sanghyun and Iyer, Adithya Jairam Vedagiri and Akula, Sai Charitha and Yang, Shusheng and Yang, Jihan and Middepogu, Manoj and Wang, Ziteng and others},
  journal={Advances in Neural Information Processing Systems},
  volume={37},
  pages={87310--87356},
  year={2024}
}

@misc{grok15v2024,
  author = {{xAI}},
  title = {Grok-1.5V Preview},
  howpublished = {\url{https://x.ai/news/grok-1.5v}},
  year = {2024},
}

@article{erqa,
  title={Gemini robotics: Bringing ai into the physical world},
  author={Team, Gemini Robotics and Abeyruwan, Saminda and Ainslie, Joshua and Alayrac, Jean-Baptiste and Arenas, Montserrat Gonzalez and Armstrong, Travis and Balakrishna, Ashwin and Baruch, Robert and Bauza, Maria and Blokzijl, Michiel and others},
  journal={arXiv preprint arXiv:2503.20020},
  year={2025}
}

@article{mmstar,
  title={Are we on the right way for evaluating large vision-language models?},
  author={Chen, Lin and Li, Jinsong and Dong, Xiaoyi and Zhang, Pan and Zang, Yuhang and Chen, Zehui and Duan, Haodong and Wang, Jiaqi and Qiao, Yu and Lin, Dahua and others},
  journal={Advances in Neural Information Processing Systems},
  volume={37},
  pages={27056--27087},
  year={2024}
}

@inproceedings{mmbench,
  title={Mmbench: Is your multi-modal model an all-around player?},
  author={Liu, Yuan and Duan, Haodong and Zhang, Yuanhan and Li, Bo and Zhang, Songyang and Zhao, Wangbo and Yuan, Yike and Wang, Jiaqi and He, Conghui and Liu, Ziwei and others},
  booktitle={European conference on computer vision},
  pages={216--233},
  year={2024},
  organization={Springer}
}

@inproceedings{mmmu,
  title={Mmmu: A massive multi-discipline multimodal understanding and reasoning benchmark for expert agi},
  author={Yue, Xiang and Ni, Yuansheng and Zhang, Kai and Zheng, Tianyu and Liu, Ruoqi and Zhang, Ge and Stevens, Samuel and Jiang, Dongfu and Ren, Weiming and Sun, Yuxuan and others},
  booktitle={Proceedings of the IEEE/CVF conference on computer vision and pattern recognition},
  pages={9556--9567},
  year={2024}
}

@article{gemini2.5,
  title={Gemini 2.5: Pushing the frontier with advanced reasoning, multimodality, long context, and next generation agentic capabilities},
  author={Comanici, Gheorghe and Bieber, Eric and Schaekermann, Mike and Pasupat, Ice and Sachdeva, Noveen and Dhillon, Inderjit and Blistein, Marcel and Ram, Ori and Zhang, Dan and Rosen, Evan and others},
  journal={arXiv preprint arXiv:2507.06261},
  year={2025}
}

@article{mimo-vl,
  title={Xiaomi MiMo-VL-Miloco Technical Report},
  author={Li, Jiaze and Chen, Jingyang and Qu, Yuxun and Xu, Shijie and Lin, Zhenru and Zhu, Junyou and Xu, Boshen and Tan, Wenhui and Fu, Pei and Ju, Jianzhong and others},
  journal={arXiv preprint arXiv:2512.17436},
  year={2025}
}

@inproceedings{siglip,
  title={Sigmoid loss for language image pre-training},
  author={Zhai, Xiaohua and Mustafa, Basil and Kolesnikov, Alexander and Beyer, Lucas},
  booktitle={Proceedings of the IEEE/CVF international conference on computer vision},
  pages={11975--11986},
  year={2023}
}

@article{deepseek-vl2,
  title={Deepseek-vl2: Mixture-of-experts vision-language models for advanced multimodal understanding},
  author={Wu, Zhiyu and Chen, Xiaokang and Pan, Zizheng and Liu, Xingchao and Liu, Wen and Dai, Damai and Gao, Huazuo and Ma, Yiyang and Wu, Chengyue and Wang, Bingxuan and others},
  journal={arXiv preprint arXiv:2412.10302},
  year={2024}
}

@book{hartley2003multiple,
  title={Multiple view geometry in computer vision},
  author={Hartley, Richard and Zisserman, Andrew},
  year={2003},
  publisher={Cambridge university press}
}

@article{seed1.5,
  title={Seed1. 5-vl technical report},
  author={Guo, Dong and Wu, Faming and Zhu, Feida and Leng, Fuxing and Shi, Guang and Chen, Haobin and Fan, Haoqi and Wang, Jian and Jiang, Jianyu and Wang, Jiawei and others},
  journal={arXiv preprint arXiv:2505.07062},
  year={2025}
}

@techreport{seed1.8,
  title={Seed1. 8 model card: Towards generalized real-world agency},
  author={Seed, Bytedance},
  year={2025},
  institution={Technical report (model card), December 2025. URL https://lf3-static~…}
}

@article{gpt4,
  title={Gpt-4 technical report},
  author={Achiam, Josh and Adler, Steven and Agarwal, Sandhini and Ahmad, Lama and Akkaya, Ilge and Aleman, Florencia Leoni and Almeida, Diogo and Altenschmidt, Janko and Altman, Sam and Anadkat, Shyamal and others},
  journal={arXiv preprint arXiv:2303.08774},
  year={2023}
}

@article{team2025kimi-vl,
  title={Kimi-vl technical report},
  author={Team, Kimi and Du, Angang and Yin, Bohong and Xing, Bowei and Qu, Bowen and Wang, Bowen and Chen, Cheng and Zhang, Chenlin and Du, Chenzhuang and Wei, Chu and others},
  journal={arXiv preprint arXiv:2504.07491},
  year={2025}
}

@article{kimi,
  title={Kimi k2: Open agentic intelligence},
  author={Team, Kimi and Bai, Yifan and Bao, Yiping and Chen, Guanduo and Chen, Jiahao and Chen, Ningxin and Chen, Ruijue and Chen, Yanru and Chen, Yuankun and Chen, Yutian and others},
  journal={arXiv preprint arXiv:2507.20534},
  year={2025}
}

@inproceedings{dynscene,
  title={Dynscene: Scalable generation of dynamic robotic manipulation scenes for embodied ai},
  author={Lee, Sangmin and Park, Sungyong and Kim, Heewon},
  booktitle={Proceedings of the IEEE/CVF Conference on Computer Vision and Pattern Recognition},
  pages={12166--12175},
  year={2025}
}

@inproceedings{bip3d,
  title={Bip3d: Bridging 2d images and 3d perception for embodied intelligence},
  author={Lin, Xuewu and Lin, Tianwei and Huang, Lichao and Xie, Hongyu and Su, Zhizhong},
  booktitle={Proceedings of the Computer Vision and Pattern Recognition Conference},
  pages={9007--9016},
  year={2025}
}

@article{rt,
  title={Rt-1: Robotics transformer for real-world control at scale},
  author={Brohan, Anthony and Brown, Noah and Carbajal, Justice and Chebotar, Yevgen and Dabis, Joseph and Finn, Chelsea and Gopalakrishnan, Keerthana and Hausman, Karol and Herzog, Alex and Hsu, Jasmine and others},
  journal={arXiv preprint arXiv:2212.06817},
  year={2022}
}

@inproceedings{shridhar2023perceiver,
  title={Perceiver-actor: A multi-task transformer for robotic manipulation},
  author={Shridhar, Mohit and Manuelli, Lucas and Fox, Dieter},
  booktitle={Conference on Robot Learning},
  pages={785--799},
  year={2023},
  organization={PMLR}
}

@inproceedings{goyal2023rvt,
  title={Rvt: Robotic view transformer for 3d object manipulation},
  author={Goyal, Ankit and Xu, Jie and Guo, Yijie and Blukis, Valts and Chao, Yu-Wei and Fox, Dieter},
  booktitle={Conference on Robot Learning},
  pages={694--710},
  year={2023},
  organization={PMLR}
}

@inproceedings{mmspatial,
  title={Mm-spatial: Exploring 3d spatial understanding in multimodal llms},
  author={Daxberger, Erik and Wenzel, Nina and Griffiths, David and Gang, Haiming and Lazarow, Justin and Kohavi, Gefen and Kang, Kai and Eichner, Marcin and Yang, Yinfei and Dehghan, Afshin and others},
  booktitle={Proceedings of the IEEE/CVF International Conference on Computer Vision},
  pages={7395--7408},
  year={2025}
}

@article{osworld,
  title={Osworld: Benchmarking multimodal agents for open-ended tasks in real computer environments},
  author={Xie, Tianbao and Zhang, Danyang and Chen, Jixuan and Li, Xiaochuan and Zhao, Siheng and Cao, Ruisheng and Hua, Toh J and Cheng, Zhoujun and Shin, Dongchan and Lei, Fangyu and others},
  journal={Advances in Neural Information Processing Systems},
  volume={37},
  pages={52040--52094},
  year={2024}
}

@article{multi-spatialmllm,
  title={Multi-spatialmllm: Multi-frame spatial understanding with multi-modal large language models},
  author={Xu, Runsen and Wang, Weiyao and Tang, Hao and Chen, Xingyu and Wang, Xiaodong and Chu, Fu-Jen and Lin, Dahua and Feiszli, Matt and Liang, Kevin J},
  journal={arXiv preprint arXiv:2505.17015},
  year={2025}
}

@article{long2026spatialreward,
  title={SpatialReward: Bridging the Perception Gap in Online RL for Image Editing via Explicit Spatial Reasoning},
  author={Long, Yancheng and Yang, Yankai and Wei, Hongyang and Chen, Wei and Zhang, Tianke and Liu, Changyi and Jiang, Kaiyu and Chen, Jiankang and Tang, Kaiyu and Wen, Bin and others},
  journal={arXiv preprint arXiv:2602.07458},
  year={2026}
}
\end{document}